\documentclass[11pt]{article}

\usepackage{microtype}
\usepackage{graphicx}
\usepackage{booktabs}
\usepackage{ziplab-tech-report}
\usepackage{hyperref}
\usepackage{amsmath}
\usepackage{amsfonts}

\usepackage{amssymb}
\usepackage{bm}
\usepackage{enumitem}
\usepackage{multirow}
\usepackage{xspace}

\hypersetup{
  colorlinks=true,
  linkcolor=ziplabblue,
  citecolor=ziplabblue,
  urlcolor=ziplabblue,
  pdftitle={TriSplat: Simulation-Ready Feed-Forward 3D Scene Reconstruction},
  pdfauthor={Weijie Wang, Zimu Li, Jinchuan Shi, Zeyu Zhang, Botao Ye, Marc Pollefeys, Donny Y. Chen, Bohan Zhuang},
  pdfsubject={arXiv preprint},
  pdfkeywords={Feed-Forward 3D Reconstruction, Triangle Splatting, Novel-View Synthesis, Surface Reconstruction}
}

\newcommand{\method}{TriSplat\xspace}
\newcommand{\boldstart}[1]{\vspace{0.1in}\noindent\textbf{#1}}
\newcommand{\trkeywords}{Feed-Forward 3D Reconstruction, Triangle Splatting, Novel-View Synthesis, Surface Reconstruction}
\newcommand{\blfootnote}[1]{%
  \begingroup
  \renewcommand\thefootnote{}\footnotetext{#1}%
  \addtocounter{footnote}{-1}%
  \endgroup
}
\techreportshorttitle{TriSplat}

\begin{document}
\thispagestyle{empty}
\vspace*{-0.78cm}


\vspace{0.42em}
\begin{center}
{\resizebox{0.98\textwidth}{!}{{\ziplabtitlefont\fontsize{18}{21}\selectfont\color{ziplabdark}
\method: Simulation-Ready Feed-Forward 3D Scene Reconstruction}}\par}
\vspace{1.20em}

{\normalsize\rmfamily\color{ziplabdark}
Weijie Wang$^{1,*}$ \hspace{0.9em}
Zimu Li$^{1,*}$ \hspace{0.9em}
Jinchuan Shi$^{1}$ \hspace{0.9em}
Zeyu Zhang$^{1}$ \hspace{0.9em}
Botao Ye$^{2,3}$\\[-0.1em]
Marc Pollefeys$^{2,4}$ \hspace{0.9em}
Donny Y. Chen$^{5}$ \hspace{0.9em}
Bohan Zhuang$^{1}$\par
}
\vspace{0.55em}
{\footnotesize\rmfamily\color{ziplabgray}
$^{1}$ Zhejiang University \quad
$^{2}$ ETH Zurich \quad
$^{3}$ ETH AI Center \quad
$^{4}$ Microsoft \quad
$^{5}$ Monash University\par
}
\end{center}
\blfootnote{$^*$ Equal contribution.}
\vspace{-0.05cm}

\begin{center}
\begin{minipage}{\textwidth}
  \centering
  \includegraphics[width=\textwidth]{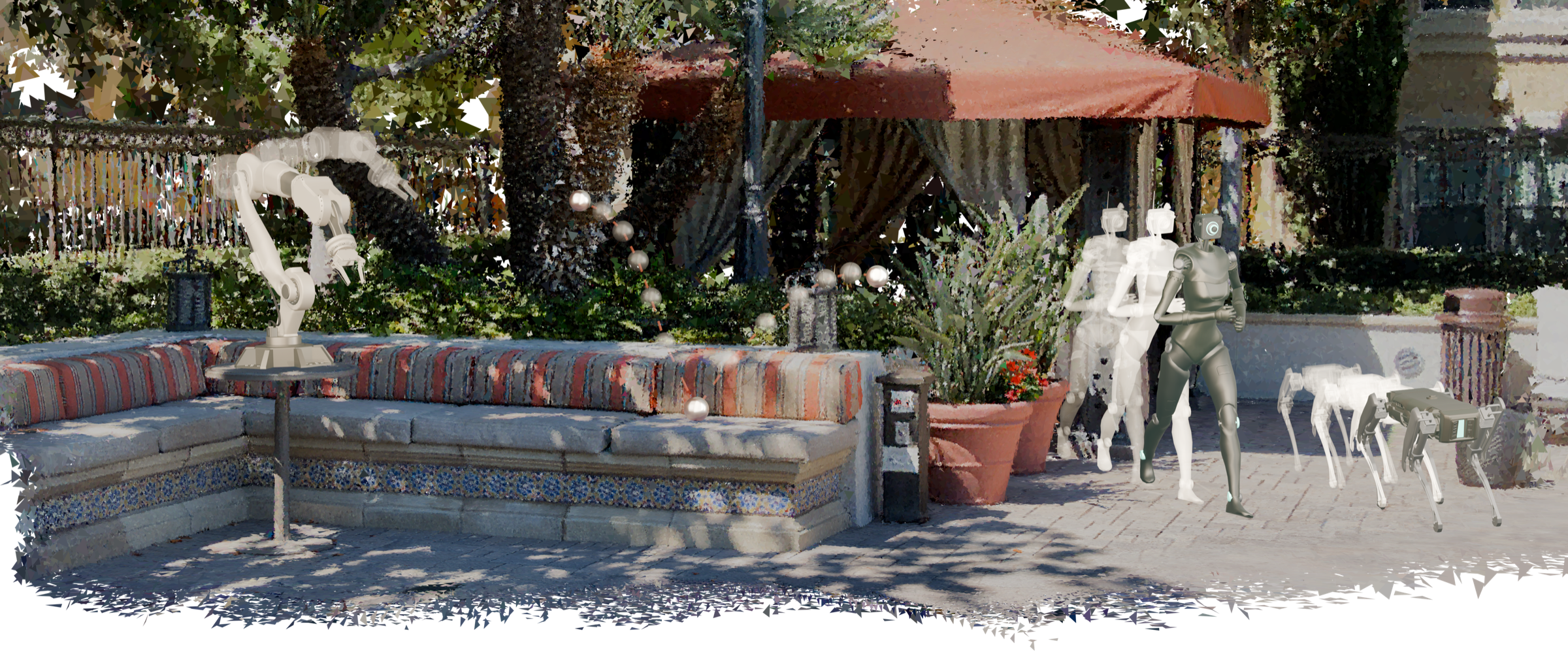}
  \vspace{-0.9em}
  {\captionsetup{font=footnotesize,aboveskip=0pt,belowskip=0pt}
  \captionof{figure}{\textbf{\method reconstructs simulation-ready 3D scenes from sparse, unposed images.} The feed-forward model predicts a triangle mesh in a single pass, enabling direct use in physics engines for locomotion, dynamics, and robotic grasping. The teaser is rendered with Blender.}}
  \label{fig:teaser}
\end{minipage}
\end{center}

\vspace{0.9em}
\begin{ziplabtitlebox}
\setlength{\parindent}{0cm}
\setlength{\parskip}{0.06cm}
\raggedright
\nohyphens

\footnotesize
\setlength{\parindent}{0cm}
\setlength{\parskip}{0.04cm}
\textbf{Abstract.}
Sparse-view 3D reconstruction is increasingly addressed with feed-forward splatting networks that predict explicit primitives directly from images. Yet most existing methods remain centered on Gaussian primitives and expose surfaces only indirectly: extracting a usable mesh for downstream simulation, physics reasoning, or embodied interaction still requires expensive post-hoc steps that break the feed-forward promise. This limitation is especially pronounced in pose-free settings, where scene structure and camera parameters must be estimated jointly from sparse observations. We present \method, a feed-forward reconstruction network that represents scenes with oriented triangle primitives and directly exports simulation-ready mesh scenes from a single forward pass. Given input images, the network predicts local 3D point maps, triangle attributes, camera poses, and optional intrinsics. Rather than regressing triangle orientation as an unconstrained latent variable, our approach constructs geometry normals from the predicted point maps, refines them with an image-conditioned normal head, and converts them into stable local frames for triangle parameterization. A mono-normal bootstrap schedule further stabilizes early training, while opacity and blur scheduling progressively sharpens the learned surface representation for direct mesh extraction. Experiments on RealEstate10K and DL3DV show that this representation produces more geometry-faithful reconstructions than Gaussian feed-forward baselines while maintaining competitive novel-view rendering quality. Because the rendering primitives are themselves surface triangles, the output can be directly ingested by physics engines, collision detectors, and standard rendering pipelines without any conversion, making it a practical simulation-ready solution for feed-forward 3D scene reconstruction.

{\setlength{\parskip}{0.1cm}\footnotesize
{\ziplabmetalabel{Project Page}\href{https://lhmd.top/trisplat}{lhmd.top/trisplat}\par}
{\ziplabmetalabel{Keywords}\trkeywords\par}
{\ziplabmetalabel{Date}May 20, 2026\par}
}
\end{ziplabtitlebox}

\clearpage
\section{Introduction}

Reconstructing 3D scenes from images is a long-standing problem in computer vision.
For robotics, augmented reality, and embodied perception~\citep{qureshi2024splatsim,chhablani2025embodiedsplat}, reconstructed scenes must support collision checking, contact-rich planning, and physics simulation.
Since engines such as NVIDIA Isaac Sim, Unity, and Unreal, as well as finite-element solvers and path tracers, build on triangle meshes, \emph{simulation-ready} reconstruction must produce explicit meshes that these engines can ingest directly.
Classical and learned multi-view pipelines~\citep{schoenberger2016sfm,schoenberger2016mvs,furukawa2015multi} can yield meshes, but they rely on multi-stage optimization, are sensitive to calibration, and degrade when views are sparse or poses are unknown.

Recent feed-forward models~\citep{yu2021pixelnerf,chen2021mvsnerf,chen2024mvsplat,zhang2025advances} sidestep per-scene optimization by predicting geometry and rendering primitives directly from images.
Gaussian splatting methods~\citep{kerbl20233d,charatan2024pixelsplat,chen2024mvsplat,xu2024depthsplat,wang2025volsplat} demonstrate efficient, high-quality novel-view synthesis, and pose-free models~\citep{wang2024dust3r,zhang2025flare,wang2025vggt,ye2024no,ye2025yonosplat} show that camera estimation and reconstruction can be learned jointly.
However, they adopt Gaussian primitives with only implicit surfaces, or point maps with no surface structure. Extracting a usable mesh then requires costly post-hoc TSDF fusion or Poisson reconstruction, breaking the feed-forward promise.
Geometry-aware variants~\citep{huang20242d,yu2024gaussian,lyu20243dgsr,chang2025meshsplat,daisurfelsplat} encourage stronger geometric structure but still rely on per-scene optimization or auxiliary extraction for mesh recovery.
On the mesh-generation side, models such as InstantMesh~\citep{xu2024instantmesh}, MeshLRM~\citep{wei2024meshlrm}, MeshFormer~\citep{liu2024meshformer}, and earlier object reconstruction methods~\citep{choy20163d,kanazawa2018learning,wang2018pixel2mesh} directly predict meshes, yet they target object-level reconstruction from controlled viewpoints and do not handle unposed, scene-level inputs.

\begin{figure}
    \centering
    \includegraphics[width=\linewidth]{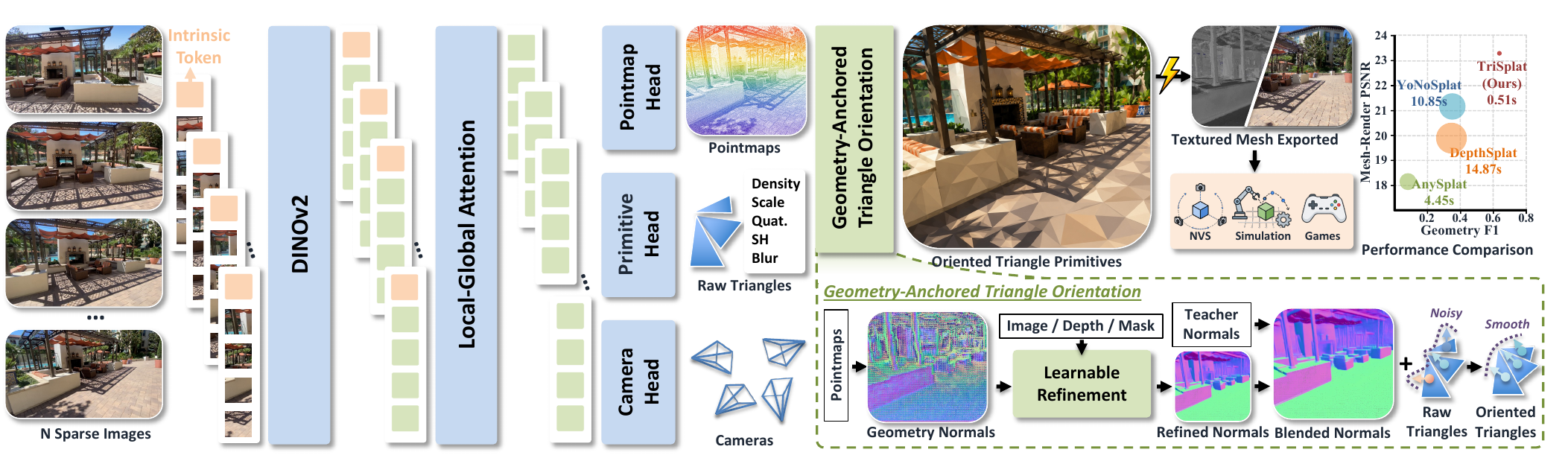}
    \vspace{-0.7cm}
    \caption{\textbf{Overview of \method.} Given $N$ sparse input images and a learnable intrinsic token, a DINOv2~\citep{oquab2023dinov2} backbone followed by Local-Global Attention decoder blocks feeds three parallel heads that predict point maps, camera poses, and raw triangle attributes (density, scale, quaternion, spherical harmonics, and blur). The dashed inset (Sec.~\ref{sec:normal_pipeline}) details the geometry-anchored triangle orientation: finite-difference geometry normals are refined by an image-conditioned U-Net, optionally blended with monocular teacher normals~\citep{eftekhar2021omnidata} in early training, and converted into tangent frames that turn noisy unoriented primitives into smooth surface elements. The resulting oriented triangles are rendered by a differentiable rasterizer~\citep{Held2025Triangle} and directly exported as a textured mesh that is immediately consumable by novel-view synthesis, physics simulation, and game engines. Top-right inset: on RealEstate10K (6 views), mesh-rendering PSNR versus mesh geometry F1, with bubble area proportional to the same-setting end-to-end runtime; up-and-right indicates better quality and a smaller bubble indicates faster inference.}
    \label{fig:pipeline}
\end{figure}

To close this gap, we present \method, a simulation-ready feed-forward model whose native representation is a set of oriented triangle primitives.
Our design follows three observations:
(i)~for simulation readiness, the rendering primitive itself must be a surface element---triangles satisfy this by construction and can be exported as a mesh without any intermediate extraction;
(ii)~triangle orientation should be \emph{anchored} to predicted local geometry rather than learned as an unconstrained variable, providing a strong prior that improves surface fidelity;
and (iii)~triangles are more sensitive to orientation errors than Gaussian splats, making explicit normal bootstrapping and validity-aware training essential.
As illustrated in Fig.~\ref{fig:pipeline}, given unposed images, \method jointly predicts local 3D point maps, per-pixel triangle attributes, camera poses, and optional focal lengths in a single forward pass.
Geometry normals from the predicted point maps are refined by an image-conditioned normal head, warm-started from a monocular teacher, and stabilized by validity-aware masking.
The refined normals form local tangent frames that orient each triangle, tying surface geometry to rendering per pixel.
Each primitive is instantiated from a canonical triangle template with learned center, scale, rotation, appearance, opacity, and blur, rendered with a differentiable triangle rasterizer~\citep{Held2025Triangle}, and sharpened from soft primitives into crisp surface elements.
Because the representation is explicitly triangular, the rendering primitives themselves form a mesh that can be loaded into physics engines, collision detectors, and standard rendering pipelines without post-processing.

Experiments on RealEstate10K~\citep{zhou2018stereo} and DL3DV~\citep{ling2024dl3dv} show that \method delivers mesh-rendering quality that surpasses state-of-the-art Gaussian feed-forward baselines while consistently outperforming them on surface accuracy metrics.
Notably, when all methods export meshes for standard triangle rendering, Gaussian baselines suffer a substantial quality drop due to lossy TSDF fusion, whereas \method exhibits minimal degradation since its rendering primitives are already the mesh.
Zero-shot evaluation on ScanNet~\citep{dai2017scannet} further confirms cross-dataset generalization, and ablation studies validate the complementary contributions of each proposed component.

Our contributions can be summarized as follows.
First, we propose \method, a feed-forward network whose native representation is oriented triangle primitives, jointly predicting geometry, appearance, and camera poses from sparse, unposed images in a single forward pass.
Second, we design a normal-anchored triangle construction pipeline that derives orientation from predicted point-map geometry, refines it with a dedicated image-conditioned head, and stabilizes training through mono-normal bootstrapping and validity-aware masking.
Third, we show that the triangle-native representation eliminates post-hoc mesh extraction: the rendering output is directly consumable by physics engines and standard rendering pipelines, making feed-forward reconstruction simulation-ready.

\section{Related Work}

\boldstart{Splatting-Based Scene Representations.}
3D Gaussian Splatting (3DGS)~\citep{kerbl20233d} represents scenes as sets of anisotropic Gaussian primitives rendered via differentiable alpha-blending, achieving real-time, high-quality novel-view synthesis. Extensions improve appearance, structure, efficiency, or compression~\citep{zhang2024fregs,jiang2024gaussianshader,lu2024scaffold,peng2024bags,zhao2024bad,lee2024compact,chen2024hac,fan2024lightgaussian,niedermayr2024compressed}, but the volumetric nature of 3D Gaussians still leads to view-inconsistent depth and poorly defined surfaces. 2DGS~\citep{huang20242d} addresses this by collapsing each Gaussian to a planar disk, producing view-consistent depth suitable for TSDF-based mesh extraction. Gaussian Opacity Fields~\citep{yu2024gaussian}, 3DGSR~\citep{lyu20243dgsr}, SurfaceSplat~\citep{gao2025surfacesplat}, and related geometry-aware 3DGS variants~\citep{fan2024trim,wolf2024surface,cheng2024gaussianpro,ververas2024sags} instead couple Gaussians with implicit, stereo, or surface fields for marching-cubes-style surface recovery. While these variants improve geometric quality, the underlying primitives remain Gaussian and meshes must be extracted through auxiliary post-processing. Triangle Splatting~\citep{Held2025Triangle} takes a fundamentally different direction by replacing Gaussians with oriented triangle primitives rendered through a differentiable rasterizer, producing an immediately exportable mesh. This validates triangle-based differentiable rendering as a viable alternative, but operates exclusively in a per-scene optimization setting.

\boldstart{Feed-Forward Sparse-View Reconstruction.}
Feed-forward methods learn scene priors from large-scale data to predict 3D representations in a single forward pass. Early image-based and NeRF-based approaches~\citep{yu2021pixelnerf,su2024boostmvsnerfs} regress radiance fields from few images but inherit costly volumetric rendering. With 3DGS, explicit feed-forward methods~\citep{wang2026feed,charatan2024pixelsplat,chen2024mvsplat,xu2024depthsplat,wewer2024latentsplat,zhang2025transplat,min2024epipolarfree,tang2024hisplat,fei2024pixelgaussian,jiang2025anysplat,wang2025freesplat++,huang2025longsplat,xiao2025jointsplat,wang2025zpressor,wang2025volsplat,shi2025pmloss,wang2025drivegen3d,liu2025trace} predict per-pixel Gaussians for efficient, high-quality novel-view synthesis from sparse inputs. A parallel line of work eliminates the requirement of known camera poses: DUSt3R~\citep{wang2024dust3r}, MASt3R~\citep{leroy2024grounding,cabon2025must3r}, VGGT~\citep{wang2025vggt}, and related models~\citep{yang2025fast3r,tang2024mv,keetha2025mapanything,jang2025pow3r} predict dense geometry to jointly recover structure and relative pose, while NoPoSplat~\citep{ye2024no}, InstantSplat~\citep{fan2024instantsplat}, Splatt3R~\citep{smart2024splatt3r}, FreeSplatter~\citep{xu2024freesplatter}, RegGS~\citep{cc2025_reggs}, UFV-Splatter~\citep{fujimura2025ufv}, FLARE~\citep{zhang2025flare}, and YoNoSplat~\citep{ye2025yonosplat} extend pose-free prediction directly to Gaussian primitives. Despite substantial progress, all these methods output Gaussians or point maps whose surface topology is only implicit.

\boldstart{Surface-Aware Feed-Forward Reconstruction.}
Recent efforts aim to combine the efficiency of feed-forward prediction with stronger surface representations. MeshSplat~\citep{chang2025meshsplat} predicts 2DGS through a dedicated normal prediction network supervised by a monocular normal estimator and regularizes positions via a weighted Chamfer distance loss, substantially improving mesh quality over baselines. SurfelSplat~\citep{daisurfelsplat} introduces Nyquist-guided surfel adaptation for feed-forward surface reconstruction. However, both methods retain Gaussian-family primitives and still rely on TSDF fusion to obtain meshes. Our method brings the triangle primitive into feed-forward, pose-free regime, where oriented triangles used for differentiable rendering can be directly exported as a mesh without additional post-processing or per-scene tuning.

\section{Method}
\label{sec:method}

Given a sparse set of $V$ unposed images $\{\mathbf{I}_v\}_{v=1}^{V}$, \method reconstructs the scene as a collection of oriented triangle primitives in a single forward pass, jointly predicting dense local 3D point maps, per-pixel triangle attributes, camera poses, and optionally camera intrinsics. Because the rendering primitives are themselves explicit surface triangles, the output can be directly exported as a mesh without any post-processing. We first describe how the network maps images to 3D points and triangle parameters in Sec.~\ref{sec:prediction}. The predicted point maps provide the geometric foundation for anchoring triangle orientation, which we detail in Sec.~\ref{sec:normal_pipeline}. The resulting oriented triangles are sharp-edged by nature and require a progressive training curriculum, presented in Sec.~\ref{sec:progressive}. Finally, Sec.~\ref{sec:supervision} describes the training objectives and the trivial mesh extraction enabled by the triangle-native representation. An overview is shown in Fig.~\ref{fig:pipeline}.

\subsection{From Images to Triangle Primitives}
\label{sec:prediction}

The encoder builds on a DINOv2~\citep{oquab2023dinov2} backbone followed by a custom transformer decoder~\citep{ye2025yonosplat}. Decoder blocks alternate between intra-view self-attention for local spatial reasoning and cross-view joint attention for multi-view correspondence aggregation, with two-dimensional rotary position embeddings and per-pixel ray-direction embeddings providing spatial and geometric conditioning throughout.

Three parallel heads convert the decoded features into scene structure, camera parameters, and primitive attributes. The point head predicts a dense local 3D point map $\mathbf{P}\!\in\!\mathbb{R}^{H\times W\times 3}$ in the coordinate frame of each camera. For each pixel it outputs three unconstrained scalars $(u, v, z')$; the depth is recovered as $z = \exp(z')$ to ensure strict positivity, and the 3D point is
\begin{equation}
    \mathbf{p} = z \cdot (u,\; v,\; 1)^\top.
    \label{eq:point}
\end{equation}
This parameterization couples lateral position with depth through multiplication, mirroring the projective image-formation model. The camera head predicts one SE(3) camera-to-world pose per view by mean-pooling decoder tokens and regressing a translation together with a $3\!\times\!3$ matrix projected onto SO(3) via SVD orthogonalization~\citep{levinson2020analysis}. All poses are expressed relative to the first view to eliminate global gauge ambiguity, and during training we apply scheduled sampling~\citep{bengio2015scheduled} that linearly decays the probability of using the ground-truth pose to prevent distribution shift at test time. The primitive head predicts per-pixel triangle attributes consisting of a density logit, three scale logits, a quaternion, spherical-harmonic appearance coefficients, and a blur parameter. To supply this branch with direct access to appearance, the input RGB image is patch-embedded and additively fused into its features before decoding. All dense heads employ pixel-shuffle upsampling~\citep{shi2016real} to reach full spatial resolution.

The predicted point maps and camera poses together define triangle centers $\mathbf{c}$ in world space. Each triangle is instantiated from a canonical equilateral template $\mathcal{T}\!\in\!\mathbb{R}^{3\times 3}$. Three raw scale logits are mapped via sigmoid to a bounded interval and converted to world-space sizes using the predicted depth and the intrinsic-derived pixel footprint. Let $\mathbf{s}$ denote the resulting scale vector, $\mathbf{R}_n$ the tangent-frame rotation that orients the triangle along the local surface (derived in Sec.~\ref{sec:normal_pipeline}), and $\mathbf{R}_c$ the camera-to-world rotation. The $k$-th vertex is
\begin{equation}
    \mathbf{v}_k = \mathbf{R}_c \, \mathbf{R}_n \bigl(\mathcal{T}_k \odot \mathbf{s}\bigr) + \mathbf{c}, \quad k \in \{1, 2, 3\},
    \label{eq:vertex}
\end{equation}
where $\odot$ denotes element-wise multiplication. The constructed triangles are rendered by a differentiable triangle rasterizer~\citep{Held2025Triangle} via tile-based sorting and front-to-back alpha compositing, producing RGB images, depth maps, and surface normals.

The point maps produced by this stage serve a dual purpose. Beyond defining triangle centers, they also provide the geometric foundation for deriving triangle orientation, as we describe next.

\subsection{Anchoring Triangle Orientation to Geometry}
\label{sec:normal_pipeline}

Triangle primitives are far more sensitive to orientation errors than Gaussian splats. A slightly misoriented Gaussian still produces a plausible soft footprint, whereas a misoriented triangle creates hard-edged artifacts whose visibility scales directly with the angular error. Treating orientation as an unconstrained latent variable is therefore impractical. We instead \emph{anchor} it to the predicted 3D geometry through the following 
pipeline that progressively refines the orientation estimate.

\boldstart{Geometry normals.}
Given the dense point map $\mathbf{P}$ from the point head, surface normals follow from finite differences. Padded horizontal and vertical derivatives $\Delta_x$ and $\Delta_y$ yield the raw geometry normal
\begin{equation}
    \mathbf{n}_{\mathrm{geo}} = \mathrm{normalize}(\Delta_x \times \Delta_y),
    \label{eq:geo_normal}
\end{equation}
which flips toward the camera when $\mathbf{n}_{\mathrm{geo}} \cdot \mathbf{p} > 0$. Border pixels and degenerate cross products are excluded via a Boolean validity mask $\mathbf{m}$ propagated through all subsequent stages. The point map may be optionally detached from the computation graph to decouple normal refinement from point prediction, and smoothed with an average-pooling kernel to suppress high-frequency noise. An orientation-aware box filter further refines the field by weighting only neighbors whose normals agree in sign with the center pixel, preserving discontinuities at depth edges.

These geometry normals provide a strong structural prior but are inevitably noisy during early training when point maps have not converged. Two complementary mechanisms address this.

\boldstart{Learned refinement.}
A lightweight U-Net incorporates appearance and depth cues not captured by local finite differences. It takes as input the channel-wise concatenation of the raw and smoothed geometry normals, the downsampled RGB image $\mathbf{I}_v$, the predicted depth map $\mathbf{D}_v\!\in\!\mathbb{R}^{H\times W}$ (whose pixel values are the per-pixel $z$ from Eq.~\eqref{eq:point}), and the validity mask. Its output layer is initialized to zero so that the head begins as an identity mapping and gradually learns corrections. Let $\mathbf{n}_{\mathrm{sm}}$ denote the smoothed geometry normal and $f_\theta$ the refinement network. The refined normal is
\begin{equation}
    \mathbf{n}_{\mathrm{ref}} = \mathrm{normalize}\,\!\bigl(\mathbf{n}_{\mathrm{sm}} + f_\theta(\mathbf{n}_{\mathrm{geo}},\, \mathbf{n}_{\mathrm{sm}},\, \mathbf{I}_v,\, \mathbf{D}_v,\, \mathbf{m})\bigr).
    \label{eq:refine}
\end{equation}
Zero-initialization is critical for stability, as a randomly initialized head would perturb orientations before useful gradients have accumulated, disrupting triangle rendering from the start.

\boldstart{Mono-normal bootstrap.}
Even with the refinement head, the earliest stage of training presents a chicken-and-egg problem: point maps are too inaccurate for reliable normals, and the refinement network has not learned meaningful corrections. We break this deadlock with a bootstrap schedule that warm-starts orientation from a pretrained monocular normal estimator~\citep{eftekhar2021omnidata}. Teacher normals $\mathbf{n}_{\mathrm{tch}}$ are computed offline for each input view, and a time-varying coefficient $\alpha(t)$ blends them with the model normals:
\begin{equation}
    \mathbf{n}_{\mathrm{fwd}} = \mathrm{normalize}\,\!\bigl(\alpha(t)\,\mathbf{n}_{\mathrm{tch}} + (1 - \alpha(t))\,\mathbf{n}_{\mathrm{ref}}\bigr).
    \label{eq:blend}
\end{equation}
The schedule comprises three phases: a \emph{takeover} phase ($t \leq t_{\mathrm{tk}}$, $\alpha\!=\!1$) where the teacher fully determines orientation; a \emph{blending} phase ($t_{\mathrm{tk}} < t < t_{\mathrm{bl}}$) where $\alpha$ decays via a cosine schedule
\begin{equation}
    \alpha(t) = \tfrac{1}{2}\!\left(1 + \cos\!\left(\pi \cdot \frac{t - t_{\mathrm{tk}}}{t_{\mathrm{bl}} - t_{\mathrm{tk}}}\right)\right);
    \label{eq:cosine}
\end{equation}
and a \emph{release} phase ($t \geq t_{\mathrm{bl}}$, $\alpha\!=\!0$) where the model relies entirely on its own geometry. Blending is restricted to pixels where both teacher and geometry validity masks hold. Importantly, this bootstrap operates on the forward-pass representation rather than on a loss term: the teacher normal directly enters triangle construction and therefore shapes the rendered output and all downstream gradients, making it fundamentally different from a teacher-matching loss that only provides an additive optimization signal. In practice we apply both simultaneously for maximum stability.

\boldstart{Tangent frame construction.}
The blended normal $\mathbf{n}_{\mathrm{fwd}}$ is converted into a full orthonormal frame $[\mathbf{t},\, \mathbf{b},\, \mathbf{n}_{\mathrm{fwd}}]$. The tangent $\mathbf{t}$ is obtained by projecting the point-map derivative $\Delta_x$ onto the plane perpendicular to $\mathbf{n}_{\mathrm{fwd}}$ and normalizing, aligning local axis of the triangle with the dominant surface gradient direction. The bitangent follows from $\mathbf{b} = \mathbf{n}_{\mathrm{fwd}} \times \mathbf{t}$, and orthogonality is guaranteed by re-deriving $\mathbf{t} = \mathbf{b} \times \mathbf{n}_{\mathrm{fwd}}$. The resulting $3\!\times\!3$ rotation matrix serves directly as $\mathbf{R}_n$ in Eq.~\eqref{eq:vertex} at valid pixels and is additionally stored as a unit quaternion for compact representation. 

With triangle orientation now anchored to geometry, the remaining challenge is that the hard-edged nature of triangles makes early-stage training unstable when predictions are still coarse.

\subsection{Progressive Surface Sharpening}
\label{sec:progressive}

A Gaussian primitive that is slightly too large or misplaced still covers roughly the correct image region through its smooth radial falloff, receiving useful gradients. A triangle in the same situation may miss its target pixels entirely, producing zero gradients and stalling learning. We address this by scheduling two complementary softness parameters that gradually transition the representation from blurred, forgiving primitives to sharp, mesh-ready surface elements.

\boldstart{Opacity scheduling.}
The predicted density $p\!\in\![0,1]$ is first converted to opacity through a nonlinear mapping whose shape changes over training. The exponent $e(t)$ ramps linearly from $e_{\mathrm{init}}$ to $e_{\mathrm{final}}$ during warm-up. The opacity is
\begin{equation}
    o = \tfrac{1}{2}\!\left(1 - (1-p)^{e(t)} + p^{\,e(t)}\right).
    \label{eq:opacity}
\end{equation}
When $e(t)\!=\!1$ the mapping reduces to identity ($o\!=\!p$); as $e(t)$ grows, intermediate densities are pushed toward zero or one, progressively binarizing the opacity field. An additional temperature factor $\tau(t)$ further sharpens the distribution at render time: the opacity is remapped via $o \leftarrow \sigma\!\bigl(\tau(t)\cdot\mathrm{logit}(o)\bigr)$, where $\tau(t)$ increases linearly from $\tau_{\mathrm{init}}$ to $\tau_{\mathrm{final}}$.

\begin{figure}[t]
\centering
\includegraphics[width=\textwidth]{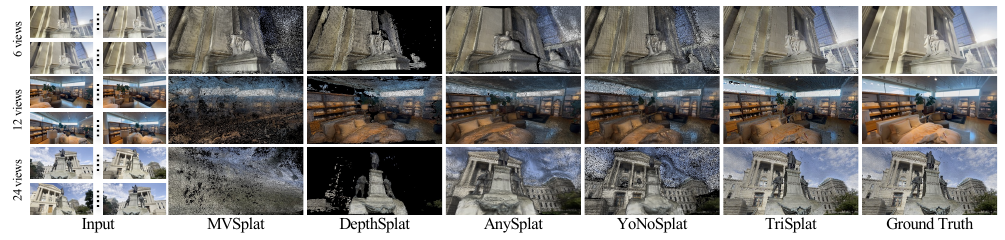}
\caption{\textbf{Mesh-rendering comparison on DL3DV.} Rows use 6, 12, and 24 input views and keep the baselines available in the source comparison. After mesh export, Gaussian baselines show missing surfaces and inconsistent geometry, whereas \method keeps more complete triangle-rendered structure.}
\label{fig:nvs_qual_mesh_dl3dv}
\end{figure}

\begin{figure}[t]
\centering
\includegraphics[width=\textwidth]{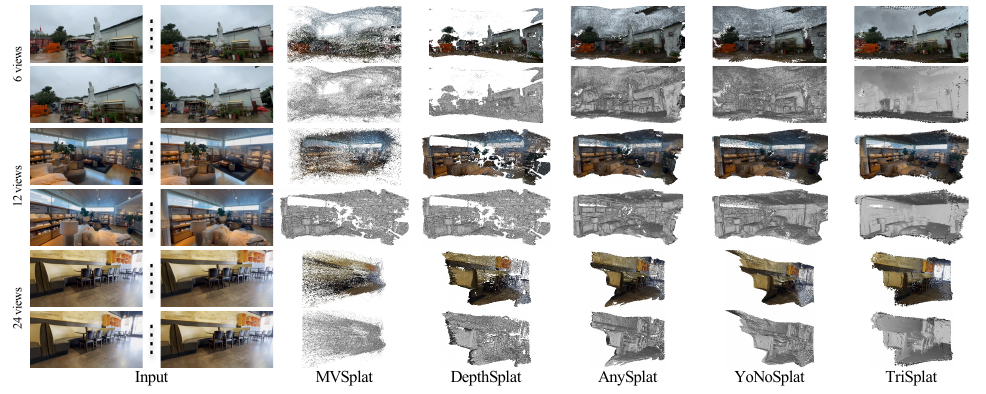}
\caption{\textbf{Textured mesh comparison on DL3DV.} Across 6/12/24 input views, Gaussian-to-TSDF conversion often shrinks or fragments geometry and loses scene extent; \method directly exports more coherent textured triangle surfaces.}
\label{fig:mesh_qual_dl3dv}
\end{figure}

\boldstart{Blur scheduling.}
Each triangle carries a scalar blur parameter modulating alpha falloff around its edges in the rasterizer:
\begin{equation}
    \sigma = \mathrm{sigmoid}(\hat{\sigma}) \cdot \beta(t),
    \label{eq:blur}
\end{equation}
where $\hat{\sigma}$ is the raw predicted value and $\beta(t)$ decays linearly from $\beta_{\mathrm{init}}$ to $\beta_{\mathrm{final}}$. Large initial blur creates broad, overlapping soft footprints with dense gradient coverage. As $\beta$ decreases, each triangle tightens into a well-defined surface element.

Opacity controls how strongly each primitive contributes to the composited color, while blur controls the spatial extent of that contribution. Scheduling both in conjunction provides a richer soft-to-crisp curriculum than either alone, ensuring stable early optimization and progressively tighter surface definition as geometry and orientation converge.

\subsection{Training Objectives and Mesh Extraction}
\label{sec:supervision}

\boldstart{Training objectives.}
\method is trained end-to-end with three complementary terms that supervise rendering, camera, and surface orientation, respectively:
\begin{equation}
    \mathcal{L} = \mathcal{L}_{\mathrm{photo}} + \mathcal{L}_{\mathrm{cam}} + \mathcal{L}_{\mathrm{normal}}.
    \label{eq:total}
\end{equation}
The photometric term $\mathcal{L}_{\mathrm{photo}}$ combines a pixel-wise reconstruction loss with a perceptual LPIPS loss~\citep{zhang2018unreasonable} between the rendered and ground-truth images. The camera term $\mathcal{L}_{\mathrm{cam}}$ is a pairwise relative pose loss over all ordered view pairs, with a Huber term on relative translations and an angular term on relative rotations; this pairwise form is invariant to the global coordinate frame and provides denser supervision than per-view absolute regression. The normal term $\mathcal{L}_{\mathrm{normal}}$ is a cosine similarity loss that aligns the refined normal $\mathbf{n}_{\mathrm{ref}}$ with the monocular teacher normal $\mathbf{n}_{\mathrm{tch}}$ at pixels where both are valid. The exact formulation of each term, the per-term loss weights, and a large-loss filter that suppresses outlier samples after warm-up are reported in Appendix~\ref{sec:app_loss}.

\begin{table}[t]
\centering
\caption{\textbf{Surface quality on DL3DV with 6/12/24 input views.} We evaluate exported meshes against the ground-truth surface. Gaussian baselines use TSDF-fused meshes, while the triangle-native model exports its primitives directly. Best results are in \textbf{bold}, second-best are \underline{underlined}.}
\label{tab:main_dl3dv_surface}
{\footnotesize
\setlength{\tabcolsep}{4.0pt}
\renewcommand{\arraystretch}{0.94}
\begin{tabular*}{0.96\textwidth}{@{\extracolsep{\fill}}l|cccc|cccc|cccc@{}}
\toprule
& \multicolumn{4}{c|}{6 views} & \multicolumn{4}{c|}{12 views} & \multicolumn{4}{c}{24 views} \\
Method & CD\,$\downarrow$ & Prec.\,$\uparrow$ & Rec.\,$\uparrow$ & F1\,$\uparrow$ & CD\,$\downarrow$ & Prec.\,$\uparrow$ & Rec.\,$\uparrow$ & F1\,$\uparrow$ & CD\,$\downarrow$ & Prec.\,$\uparrow$ & Rec.\,$\uparrow$ & F1\,$\uparrow$ \\
\midrule
MVSplat
& 1.143 & 0.121 & 0.154 & 0.118
& 0.802 & 0.122 & 0.211 & 0.135
& 0.695 & \underline{0.130} & \underline{0.276} & \underline{0.156} \\
DepthSplat
& 1.116 & \underline{0.135} & \underline{0.192} & \underline{0.145}
& 0.907 & \underline{0.127} & \underline{0.237} & \underline{0.152}
& 0.786 & 0.120 & 0.274 & 0.152 \\
AnySplat
& 1.012 & 0.088 & 0.143 & 0.093
& 0.731 & 0.085 & 0.163 & 0.096
& 0.699 & 0.084 & 0.175 & 0.100 \\
YoNoSplat
& \underline{0.920} & 0.090 & 0.189 & 0.106
& \underline{0.664} & 0.073 & 0.183 & 0.092
& \underline{0.687} & 0.068 & 0.173 & 0.088 \\
\midrule
\method (Ours)
& \textbf{0.613} & \textbf{0.223} & \textbf{0.448} & \textbf{0.287}
& \textbf{0.323} & \textbf{0.200} & \textbf{0.522} & \textbf{0.279}
& \textbf{0.310} & \textbf{0.191} & \textbf{0.571} & \textbf{0.277} \\
\bottomrule
\end{tabular*}
}
\end{table}

\begin{table}[t]
\centering
\caption{\textbf{NVS quality on DL3DV with 6/12/24 input views under mesh rendering.} We render each exported mesh with the same triangle rasterizer and report target-view reconstruction quality.}
\label{tab:main_dl3dv_nvs}
{\footnotesize
\setlength{\tabcolsep}{3.8pt}
\renewcommand{\arraystretch}{0.94}
\begin{tabular*}{0.96\textwidth}{@{\extracolsep{\fill}}l|ccc|ccc|ccc@{}}
\toprule
& \multicolumn{3}{c|}{6 views} & \multicolumn{3}{c|}{12 views} & \multicolumn{3}{c}{24 views} \\
Method & PSNR\,$\uparrow$ & SSIM\,$\uparrow$ & LPIPS\,$\downarrow$ & PSNR\,$\uparrow$ & SSIM\,$\uparrow$ & LPIPS\,$\downarrow$ & PSNR\,$\uparrow$ & SSIM\,$\uparrow$ & LPIPS\,$\downarrow$ \\
\midrule
MVSplat    & 14.75 & 0.450 & 0.509 & 15.16 & 0.431 & 0.525 & 15.72 & 0.439 & 0.526 \\
DepthSplat & 14.86 & 0.481 & 0.490 & 14.82 & 0.426 & 0.536 & 15.13 & 0.422 & 0.541 \\
AnySplat   & 18.58 & \underline{0.551} & \underline{0.397} & 16.58 & \underline{0.485} & \underline{0.462} & 16.42 & 0.442 & \underline{0.475} \\
YoNoSplat  & \underline{18.88} & 0.548 & 0.414 & \underline{16.90} & 0.459 & 0.485 & \underline{16.71} & \underline{0.452} & 0.499 \\
\midrule
\method (Ours) & \textbf{20.84} & \textbf{0.615} & \textbf{0.335} & \textbf{18.71} & \textbf{0.493} & \textbf{0.400} & \textbf{18.06} & \textbf{0.455} & \textbf{0.425} \\
\bottomrule
\end{tabular*}
}
\end{table}

\boldstart{Mesh extraction.}
A distinctive advantage of the triangle-native representation is that mesh extraction becomes trivial. Because the rendering output already consists of oriented triangles in world space, no auxiliary reconstruction is needed. After a forward pass, low-opacity triangles are discarded, winding order is corrected by comparing face normals against per-pixel normals from Sec.~\ref{sec:normal_pipeline}, and nearby duplicate vertices are merged via quantized position hashing. The result is a standard triangle mesh produced without per-scene optimization, TSDF fusion, or marching cubes, directly usable in physics simulation, collision detection, and standard rendering engines.

\section{Experiments}
\label{sec:experiments}

We evaluate \method along three axes that directly reflect its simulation-ready objective: (i) the quality of the reconstructed surface geometry, (ii) novel-view rendering quality when the exported mesh is consumed by a standard triangle rasterizer, (iii) depth and normal accuracy, and (iv) runtime efficiency. All design choices are additionally validated through controlled ablation studies.

\subsection{Experimental Setup}
\label{sec:setup}

\boldstart{Datasets.}
We train on RealEstate10K (RE10K)~\citep{zhou2018stereo} and DL3DV~\citep{ling2024dl3dv} following standard splits~\cite{ye2025yonosplat}. RE10K contains 67{,}477 training and 7{,}289 test scenes collected from real-estate walkthroughs on YouTube, spanning diverse indoor and outdoor environments with camera parameters recovered via structure-from-motion. DL3DV contains over 10{,}000 real-world scenes captured at high resolution, offering richer complexity and wider viewpoint variation than prior datasets. We additionally evaluate zero-shot generalization on 100 held-out scenes from ScanNet~\citep{dai2017scannet} following MeshSplat~\citep{chang2025meshsplat}.

\begin{figure}[t]
\centering
\includegraphics[width=\textwidth]{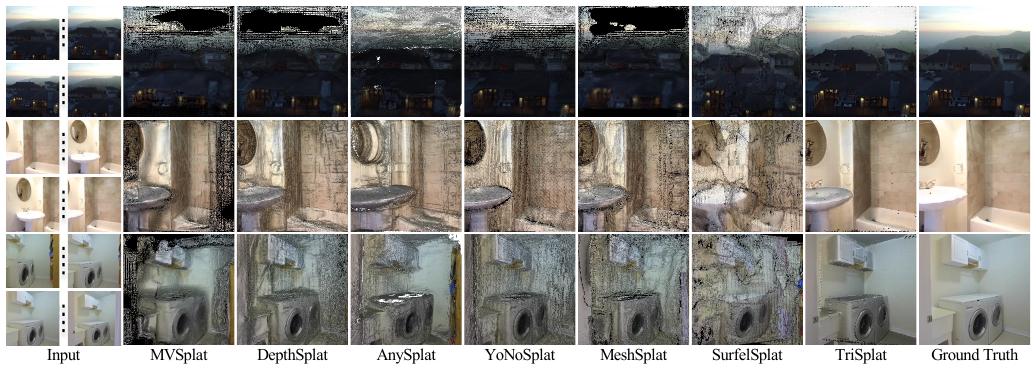}
\caption{\textbf{Mesh-rendering comparison on RE10K.} We compare target-view renders obtained from exported meshes using six input views and the same mesh-rendering protocol. TSDF-fused Gaussian baselines blur surfaces, drop structures, or introduce floaters, while \method preserves sharper triangle-rendered detail and more complete silhouettes.}
\label{fig:nvs_qual_mesh}
\label{fig:nvs_qual_mesh_re10k}
\end{figure}

\begin{figure}[t]
\centering
\includegraphics[width=\textwidth]{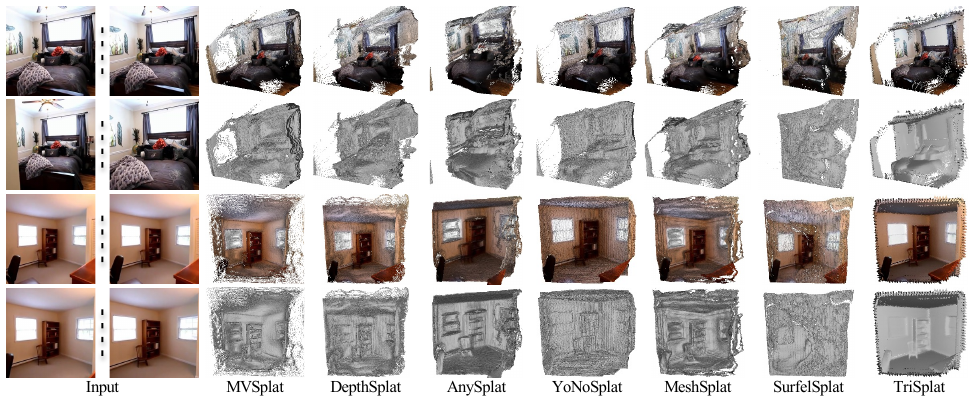}
\caption{\textbf{Textured mesh comparison on RE10K.} We visualize exported textured meshes rather than image-space renders, including both textured and geometry-only views. \method yields cleaner triangle surfaces, while TSDF-fused Gaussian baselines over-smooth or fragment geometry.}
\label{fig:mesh_qual}
\label{fig:mesh_qual_re10k}
\end{figure}

\boldstart{Baselines.}
We compare against feed-forward methods spanning Gaussian splatting and their variants: MVSplat~\citep{chen2024mvsplat} and DepthSplat~\citep{xu2024depthsplat} (cost-volume-based Gaussian models), AnySplat~\citep{jiang2025anysplat} and YoNoSplat~\citep{ye2025yonosplat} (pose-free Gaussian methods), and MeshSplat~\citep{chang2025meshsplat} and SurfelSplat~\citep{daisurfelsplat} (geometry-aware variants).

\boldstart{Metrics.}
Surface quality is measured by Chamfer Distance (CD), Precision, Recall, and F1 score, computed with the protocol detailed in Appendix~\ref{sec:app_mesh_eval}. Rendering quality is measured by PSNR, SSIM~\citep{wang2004image}, and LPIPS~\citep{zhang2018unreasonable} on mesh. For ScanNet we additionally report depth accuracy (AbsRel, AbsDiff) and normal accuracy (mean angular error, fraction of pixels within $30^\circ$).

\begin{figure}[t]
\centering
\includegraphics[width=\textwidth]{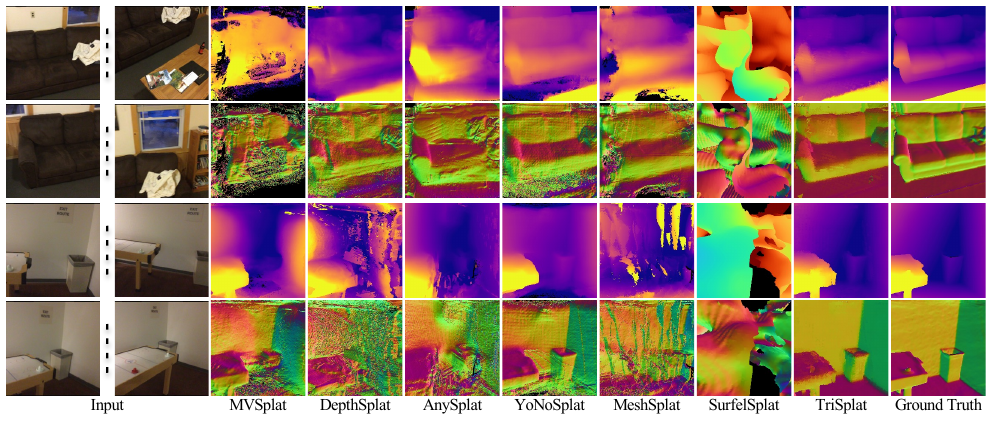}
\caption{\textbf{Depth and normal comparison on ScanNet.} All models are trained on RE10K and evaluated zero-shot on ScanNet, with depth and surface normals shown for each method. \method produces smoother surface-aligned normals and sharper depth boundaries under this domain shift.}
\label{fig:depth_normal_qual}
\end{figure}

\begin{table}[t]
\centering
\begin{minipage}[t]{0.60\textwidth}
\centering
\caption{\textbf{Quantitative comparison on RE10K with 6 input views.} Following the same mesh-based protocol as Tables~\ref{tab:main_dl3dv_surface} and~\ref{tab:main_dl3dv_nvs}, we additionally include surface-aware variants (MeshSplat~\citep{chang2025meshsplat}, SurfelSplat~\citep{daisurfelsplat}) that target geometric fidelity rather than rendering.
}
\label{tab:main_re10k}
\vspace{-0.4em}
{\scriptsize
\setlength{\tabcolsep}{1.8pt}
\renewcommand{\arraystretch}{0.92}
\begin{tabular*}{\linewidth}{@{\extracolsep{\fill}}l|cccc|ccc@{}}
\toprule
& \multicolumn{4}{c|}{Surface Quality} & \multicolumn{3}{c}{NVS Quality} \\
Method & CD\,$\downarrow$ & Prec.\,$\uparrow$ & Rec.\,$\uparrow$ & F1\,$\uparrow$ & PSNR\,$\uparrow$ & SSIM\,$\uparrow$ & LPIPS\,$\downarrow$ \\
\midrule
MVSplat       & 0.340 & 0.618 & 0.257 & 0.358 &  13.97 & 0.615 & 0.378 \\
DepthSplat    & 0.294 & 0.647 & 0.326 & 0.429 & 21.23 & \underline{0.781} & 0.271 \\
AnySplat      & 0.540 & 0.460 & 0.064 & 0.110 & 18.23 & 0.589 & 0.365 \\
YoNoSplat     & \underline{0.267} & \underline{0.681} & \underline{0.333} & \underline{0.443} &  \underline{21.94} & 0.753 & \textbf{0.238}  \\
MeshSplat     & 0.349 & 0.642 & 0.234 & 0.340 & 19.97 & 0.719 & 0.294 \\
SurfelSplat   & 0.747 & 0.432 & 0.099 & 0.154 &  11.18 & 0.184 & 0.738\\
\midrule
\method (Ours) & \textbf{0.190} & \textbf{0.708} & \textbf{0.560} & \textbf{0.622} & \textbf{24.69} & \textbf{0.798} & \underline{0.269} \\
\bottomrule
\end{tabular*}
}
\end{minipage}\hfill
\begin{minipage}[t]{0.37\textwidth}
\centering
\caption{\textbf{Zero-shot depth and normal evaluation on ScanNet with 6 input views.} All models are trained on RE10K and zero-shot evaluated on ScanNet.
}
\label{tab:depth_normal}
\vspace{-0.4em}
{\scriptsize
\setlength{\tabcolsep}{1.8pt}
\renewcommand{\arraystretch}{0.92}
\begin{tabular*}{\linewidth}{@{\extracolsep{\fill}}l|cc|cc@{}}
\toprule
& \multicolumn{2}{c|}{Depth} & \multicolumn{2}{c}{Normal} \\
Method & Rel\,$\downarrow$ & Diff\,$\downarrow$ & Mean\,$\downarrow$ & $<30^\circ$\,$\uparrow$ \\
\midrule
MVSplat       & 0.708 & 1.206 & 102.247 & 17.204 \\
DepthSplat    & 0.279 & 0.595 &  54.861 & 29.403 \\
AnySplat      & 0.453 & \textbf{0.283} &  55.557 & 25.375 \\
YoNoSplat     & \underline{0.270} & 0.516 &  \underline{54.110} & \underline{41.047} \\
MeshSplat     & 0.534 & 0.999 &  59.803 & 31.862 \\
SurfelSplat   & 0.716 & 1.264 &  75.300 & 16.484 \\
\midrule
\method (Ours) & \textbf{0.188} & \underline{0.341} & \textbf{27.901} & \textbf{71.708} \\
\bottomrule
\end{tabular*}
}
\end{minipage}
\end{table}

\boldstart{Evaluation protocol and rendering modes.}
On DL3DV we evaluate with 6, 12, and 24 input views at context gaps of 50--180 frames; on RE10K we evaluate with 6 views at gaps of 50-150 frames; ScanNet is evaluated in a zero-shot setting using RE10K-trained models without fine-tuning. For each method we consider two rendering modes. \emph{Primitive rendering} uses each method's native rasterizer (Gaussian splatting for baselines, triangle splatting for \method). \emph{Mesh rendering} rasterizes the \emph{exported mesh} with a standard triangle rasterizer: Gaussian baselines export meshes via TSDF fusion~\citep{chang2025meshsplat}, whereas \method exports its triangle primitives directly without any auxiliary reconstruction. Because the ultimate goal of \method is a simulation-ready mesh that can be ingested by physics engines and standard graphics pipelines, we adopt \emph{mesh rendering} as the primary rende ring metric throughout the main paper and report \emph{primitive rendering} results in Appendix~\ref{sec:app_prim_render} for reference, together with an explicit primitive-to-mesh degradation analysis. Implementation details are provided in Appendix~\ref{sec:app_implementation}.

\subsection{Surface Reconstruction and Mesh Rendering}
\label{sec:main_results}

Tables~\ref{tab:main_dl3dv_surface},~\ref{tab:main_dl3dv_nvs}, and~\ref{tab:main_re10k} jointly report the geometric quality of the exported mesh and the novel-view rendering quality when the same mesh is rasterized by a standard triangle pipeline. This unified view directly measures how faithful the reconstructed surface is \emph{and} how well it performs as the actual rendering primitive in downstream pipelines.

\boldstart{Surface geometry.}
\method consistently produces the most accurate surface geometry across all four metrics on both datasets. On RE10K (Table~\ref{tab:main_re10k}) \method attains a Chamfer Distance of~\textbf{0.190} and an F1 score of~\textbf{0.622}, improving over the strongest Gaussian baseline (YoNoSplat) by~\textbf{0.077} in CD and~\textbf{0.179} in F1. The gap is especially pronounced on Recall ($+$\textbf{0.227}), revealing that TSDF-fused meshes from Gaussian baselines systematically under-cover the ground-truth surface, particularly around thin structures. Dedicated surface-oriented Gaussian variants do not close this gap. MeshSplat achieves surface-like regularity but remains bounded by TSDF discretization, and SurfelSplat degrades substantially on CD and F1. The same pattern holds on DL3DV (Tables~\ref{tab:main_dl3dv_surface} and~\ref{tab:main_dl3dv_nvs}) across 6, 12, and 24 views, showing that \method's geometric advantage is robust to the density of input observations. Qualitative textured mesh visualizations in Figs.~\ref{fig:mesh_qual_re10k} and~\ref{fig:mesh_qual_dl3dv} confirm the numerical trends. TSDF-fused baselines produce bumpy surfaces with floaters and missing thin structures, while \method produces clean triangle meshes that preserve fine-scale geometry.

\boldstart{Mesh rendering.}
Because the downstream consumer of a simulation-ready representation is a standard triangle rasterizer, we evaluate rendering quality directly on the exported mesh. \method obtains the best mesh-rendering quality across datasets. On RE10K \method reaches~\textbf{24.69}~dB PSNR under mesh rendering, compared to~\textbf{21.94}~dB for the strongest Gaussian baseline, a margin of~$+$\textbf{2.75}~dB. The advantage stems from a structural asymmetry between the two families: Gaussian baselines incur a substantial quality drop when their TSDF-fused meshes are rendered as triangles, because the discretized volume discards the very primitives that produced the original image, whereas in \method the rendering primitives \emph{are} the mesh and no information is lost during export. Qualitative mesh-rendering results in Figs.~\ref{fig:nvs_qual_mesh_re10k} and~\ref{fig:nvs_qual_mesh_dl3dv} visualize this effect: \method preserves sharp edges and thin structures, whereas TSDF-based baselines exhibit blurred boundaries and missing geometry. A complementary analysis of the primitive-rendering mode, corresponding to each method's native rasterization prior to mesh export, is reported in Appendix~\ref{sec:app_prim_render}, together with an explicit primitive-to-mesh degradation summary.

\subsection{Depth and Normal Quality}
\label{sec:depth_normal}

Table~\ref{tab:depth_normal} evaluates depth and normal accuracy on ScanNet in a zero-shot setting, using RE10K-trained models without fine-tuning. \method achieves an AbsRel of~\textbf{0.188} and an AbsDiff of~\textbf{0.341}, the best among all compared methods. On normal metrics \method outperforms all baselines by a clear margin, with a mean angular error of~\textbf{27.9}$^\circ$ and a $<$30$^\circ$ accuracy of~\textbf{71.7}\%, compared to~\textbf{54.1}$^\circ$ mean error and~\textbf{41.0}\% at $<$30$^\circ$ for the strongest baseline. This improvement directly reflects the geometry-anchored normal pipeline and the bootstrap schedule, which explicitly optimize for orientation quality. Qualitative depth and normal maps in Fig.~\ref{fig:depth_normal_qual} confirm that \method yields smooth, geometrically coherent normals aligned with surface boundaries, whereas Gaussian baselines produce noisy, per-pixel-inconsistent normal fields. Additional novel-view synthesis results on ScanNet are provided in Appendix~\ref{sec:supp_scannet_nvs}.

\begin{figure}[t]
\centering
\includegraphics[width=0.8\columnwidth]{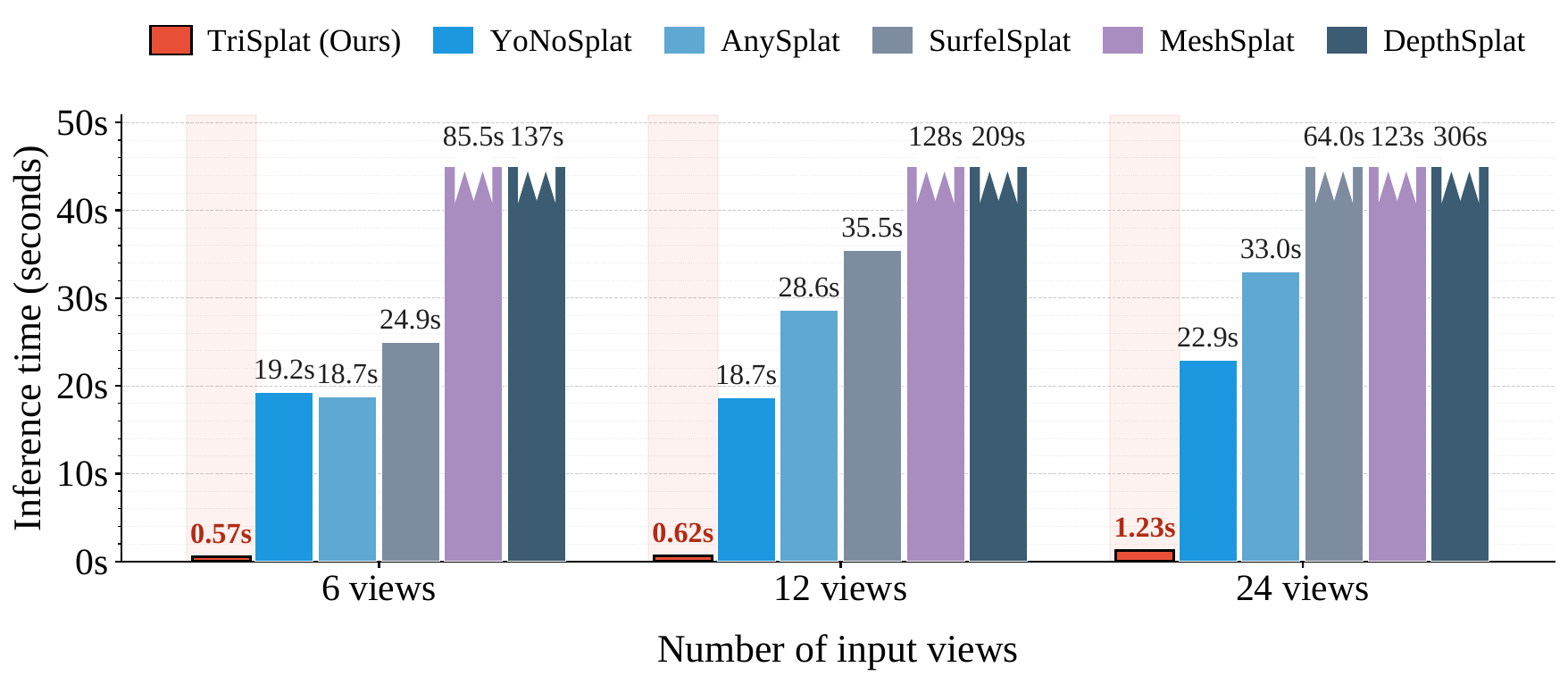}
\vspace{-0.4cm}
\caption{\textbf{Runtime comparison on DL3DV.} \method produces a usable mesh in well under~\textbf{1.3}\,s for up to 24 input views, while Gaussian feed-forward baselines additionally pay a mesh export cost that scales with the reconstructed volume. Bars beyond the 45\,s axis cap are broken; the exact value is annotated above each bar.}
\label{fig:efficiency}
\end{figure}

\subsection{Efficiency}
\label{sec:efficiency}

Fig.~\ref{fig:efficiency} compares the end-to-end time-to-mesh of all methods on DL3DV at 6, 12, and 24 input views, measured on a single NVIDIA H100 GPU. Because \method's rendering primitives are themselves the mesh, its end-to-end cost equals the feed-forward pass alone: \textbf{0.57}\,s, \textbf{0.62}\,s, and \textbf{1.23}\,s, respectively. Every Gaussian baseline, in contrast, must run an additional TSDF-fusion stage to obtain a mesh consumable by a standard triangle pipeline, and this stage scales with the reconstructed volume rather than with the network. As a result, the fastest Gaussian baseline (AnySplat) takes~\textbf{18.7}\,s at 6 views and~\textbf{33.0}\,s at 24 views, while volumetric cost-volume methods such as DepthSplat reach~\textbf{306}\,s at 24 views. End-to-end, \method is~\textbf{33}$\times$ faster than the fastest Gaussian baseline at 6 views and up to~\textbf{249}$\times$ faster than the slowest baseline at 24 views, and is the only method that remains well under one second per scene at the smallest input setting. This advantage is structural rather than incidental: eliminating the post-hoc mesh-extraction step is precisely what makes the triangle-native representation simulation-ready by design.

\subsection{Simulation-Ready Demonstration}
\label{sec:sim_demo}

\begin{figure}[t]
\centering
\includegraphics[width=\textwidth]{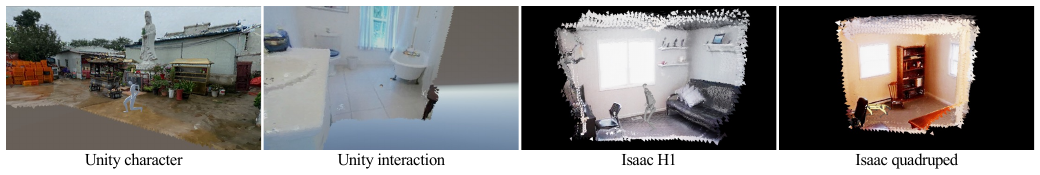}
\caption{\textbf{Simulation-ready demonstration.} Directly exported \method meshes are loaded into Unity and NVIDIA Isaac Sim for interaction, locomotion, and collision. The same mesh assets can be reused without Gaussian-to-mesh conversion or scene-specific cleanup.}
\label{fig:sim_demo}
\end{figure}

To validate the practical utility of our simulation-ready representation, we load the directly exported meshes from \method into two mainstream physics engines, Unity and NVIDIA Isaac Sim, and demonstrate a range of embodied tasks including rigid-body simulation, collision detection, robot navigation, and robotic grasping (Fig.~\ref{fig:sim_demo}). The exported meshes are consumed without any manual cleanup or format conversion. By contrast, Gaussian baselines require TSDF fusion followed by additional mesh cleaning before they can be loaded into the same engines. Extended simulation experiments are provided in Appendix~\ref{sec:app_simulation}.

\subsection{Ablation Study}
\label{sec:ablation}

We ablate the four design choices of \method on RE10K with 6 input views; Table~\ref{tab:ablation} reports surface geometry (CD, F1) and mesh-rendering quality (PSNR, LPIPS). Each component targets a distinct failure mode and removing any single one degrades all four metrics by a comparable margin. \emph{Normal anchoring} fixes orientation: replacing it with an unconstrained quaternion lets triangle centers drift, dropping F1 by~\textbf{0.057} and PSNR by~\textbf{1.11}~dB. The \emph{mono-normal bootstrap} resolves the early-stage deadlock between point maps and normals; removing it yields the largest surface degradation (F1~$-$\textbf{0.065}, PSNR~$-$\textbf{1.08}~dB). \emph{Normal refinement} suppresses finite-difference noise at depth edges that otherwise surfaces directly as rasterization artifacts, producing the largest rendering drop (PSNR~$-$\textbf{1.58}~dB, LPIPS~$+$\textbf{0.111}). \emph{Progressive sharpening} avoids the cold-start of hard-edged triangles by providing soft-footprint gradients early on; disabling it lowers PSNR by~\textbf{1.44}~dB and F1 by~\textbf{0.062} while leaving CD nearly unchanged.

\begin{table}[t]
\centering
\caption{\textbf{Ablation study on RE10K (6 views).} Each row removes one component from the full model. Geometry metrics are computed on the exported mesh; rendering metrics are reported under mesh rendering.}
\label{tab:ablation}
{\footnotesize
\setlength{\tabcolsep}{5.2pt}
\renewcommand{\arraystretch}{0.94}
\begin{tabular}{l|cc|cc}
\toprule
& \multicolumn{2}{c|}{Surface Quality} & \multicolumn{2}{c}{NVS Quality} \\
Configuration & CD $\downarrow$ & F1 $\uparrow$ & PSNR $\uparrow$ & LPIPS $\downarrow$ \\
\midrule
Full model                          & 0.190 & 0.708 & 23.25 & 0.318 \\
\midrule
w/o normal anchoring                & 0.190 & 0.651 & 22.14 & 0.396 \\
w/o mono-normal bootstrap           & 0.198 & 0.643 & 22.17 & 0.397 \\
w/o normal refinement               & 0.193 & 0.649 & 21.67 & 0.429 \\
w/o progressive sharpening          & 0.191 & 0.646 & 21.81 & 0.416 \\
\bottomrule
\end{tabular}
}
\end{table}

\section{Conclusion}
\label{sec:conclusion}

We presented \method, a feed-forward reconstruction model that represents scenes natively as oriented triangle primitives and jointly predicts geometry, appearance, and camera parameters from sparse unposed images in a single forward pass. By anchoring triangle orientation to predicted point-map geometry, warm-starting the orientation field through a mono-normal bootstrap, and bridging soft-to-crisp optimization with a progressive sharpening curriculum, \method attains substantially more accurate surface geometry than Gaussian feed-forward baselines and, because its rendering primitives are themselves the exported mesh, also yields the strongest mesh-rendering quality across RealEstate10K, DL3DV, and zero-shot ScanNet, while sidestepping the primitive-to-mesh degradation that TSDF fusion imposes on Gaussian-based pipelines. The directly exported meshes can be ingested by mainstream physics engines without any post-processing, recasting simulation readiness as a property of the representation itself rather than a downstream conversion problem.

\boldstart{Limitations and future work.}
The direct export yields a non-manifold triangle soup adequate for rendering and physics but not for applications requiring watertight meshes such as finite-element analysis, and per-pixel prediction ties triangle density to input resolution, leaving topology-aware export and adaptive tessellation as promising future directions.

\section*{Acknowledgements}

We thank ETH Zurich for providing the computational resources used in this work.

\bibliographystyle{unsrtnat}
\bibliography{reference}

\clearpage
\appendix

\setcounter{table}{0}
\setcounter{figure}{0}
\setcounter{section}{0}
\renewcommand{\thetable}{\Alph{table}}
\renewcommand{\thefigure}{\Alph{figure}}
\renewcommand{\thesection}{\Alph{section}}
\setlength{\abovecaptionskip}{-0.1em}
\setlength{\belowcaptionskip}{0pt}

\section{Implementation Details}
\label{sec:app_implementation}

\subsection{Network Architecture}
\label{sec:app_architecture}

\boldstart{Backbone.}
The encoder adopts a DINOv2 ViT-L/14~\citep{oquab2023dinov2} backbone with patch size~14 and augments it with 2D rotary position embeddings. Decoder blocks alternate between intra-view self-attention and cross-view joint attention, producing feature tokens of dimension $d\!=\!1024$. For pose-free operation the backbone additionally embeds per-pixel intrinsic information via a 4th-degree positional encoding applied at pixel level.

\boldstart{Point and camera heads.}
The point head is a 5-layer transformer decoder (dimension 1024, 16 heads, MLP ratio~4) followed by a linear projection that outputs three channels (two lateral coordinates and one log-depth), upsampled via pixel-shuffle~\citep{shi2016real} to $14\!\times$ the token resolution. The camera head shares the same transformer depth and head count but reduces the output dimension to~512, mean-pools the decoded tokens via adaptive average pooling, and maps them through two residual $1\!\times\!1$ convolution blocks and two MLP layers to produce per-view SE(3) poses.

\boldstart{Primitive head.}
The primitive head is structurally identical to the point head but its input features are additively fused with zero-initialized patch-embedded RGB tokens before decoding, providing the branch with direct access to appearance information. Its output dimension is $1 + d_{\mathrm{tri}}$, where $d_{\mathrm{tri}} = 3 + 4 + 3 + 1 = 11$ consists of three scale logits, a four-component quaternion, three zeroth-order spherical-harmonic (SH) coefficients, and one blur parameter. All dense heads employ an upscale token ratio of~2 and generate predictions at $14 \times 14$ points per patch, yielding a dense prediction map at the input image resolution.

\boldstart{Normal refinement U-Net.}
The geometry-anchored normal refinement head is a lightweight U-Net with 4 encoder--decoder scales. Each scale consists of a convolution stage (conv $\to$ GroupNorm $\to$ GELU) followed by 2 residual convolution blocks, all using $3\!\times\!3$ kernels. The channel progression through the encoder is $36 \to 72 \to 144 \to 288$ and is mirrored in the decoder, which upsamples with bilinear interpolation and concatenates skip features. The network receives 11 input channels comprising the raw geometry normal (3), the smoothed geometry normal (3), the downsampled RGB image (3), the predicted depth (1), and the validity mask (1). The output layer operates in residual mode with a scale factor of~0.25, where both weights and biases are zero-initialized so that the head starts as an identity mapping and gradually learns corrections. Training uses mixed-precision bfloat16 with gradient checkpointing enabled to reduce memory.

\subsection{Loss Formulation}
\label{sec:app_loss}

We expand each term of the training objective $\mathcal{L}$ in Sec.~\ref{sec:supervision} of the main paper. With per-term weights, the total objective is
\begin{equation}
    \mathcal{L} = \lambda_{\mathrm{photo}}\,\mathcal{L}_{\mathrm{photo}} + \lambda_{\mathrm{cam}}\,\mathcal{L}_{\mathrm{cam}} + \lambda_{\mathrm{normal}}\,\mathcal{L}_{\mathrm{normal}}.
    \label{eq:app_total}
\end{equation}

\boldstart{Photometric term.}
Let $\hat{\mathbf{I}}$ and $\mathbf{I}^{*}$ denote the rendered and ground-truth images. The photometric term combines a pixel-wise mean-squared error and a perceptual LPIPS loss~\citep{zhang2018unreasonable}:
\begin{equation}
    \mathcal{L}_{\mathrm{photo}} = \lambda_{\mathrm{mse}}\,\bigl\|\hat{\mathbf{I}} - \mathbf{I}^{*}\bigr\|_2^2 + \lambda_{\mathrm{lpips}}\,\mathrm{LPIPS}(\hat{\mathbf{I}},\,\mathbf{I}^{*}).
    \label{eq:app_photo}
\end{equation}

\boldstart{Camera term.}
The camera term is a sum over all ordered view pairs of a Huber loss $\mathcal{L}_{\mathrm{trans}}$ on the relative translation and an angular loss $\mathcal{L}_{\mathrm{rot}}$ on the relative rotation:
\begin{equation}
    \mathcal{L}_{\mathrm{cam}} = \omega_t\,\mathcal{L}_{\mathrm{trans}} + \omega_r\,\mathcal{L}_{\mathrm{rot}}.
    \label{eq:app_cam}
\end{equation}
The pairwise form is invariant to the choice of global coordinate frame and provides denser supervision than per-view absolute regression, since every view pair contributes an independent constraint.

\boldstart{Normal term.}
Let $\mathcal{V}$ be the set of valid pixels (those satisfying the geometry mask, the finite-value check, and an optional object mask). The normal term is a cosine-similarity loss between the refined normal $\mathbf{n}_{\mathrm{ref}}$ and the monocular teacher normal $\mathbf{n}_{\mathrm{tch}}$:
\begin{equation}
    \mathcal{L}_{\mathrm{normal}} = \frac{1}{|\mathcal{V}|}\sum_{i \in \mathcal{V}} \bigl(1 - \mathbf{n}_{\mathrm{ref},i}^{\top}\mathbf{n}_{\mathrm{tch},i}\bigr).
    \label{eq:app_normal}
\end{equation}

\subsection{Training Protocol}
\label{sec:app_training}

\boldstart{Pre-training initialization.}
The backbone and decoder weights are initialized from PI3~\citep{wang2025pi}, a pretrained pose-free Gaussian splatting model. The normal refinement U-Net and all triangle-specific adapter parameters are initialized from scratch with zero-initialization as described above.

\boldstart{Training schedule.}
All training and testing are conducted on NVIDIA A100 GPUs. For RE10K, we train for 150K steps at $224 \times 224$ resolution. For DL3DV, we first train for 100K steps at $224 \times 224$ resolution, then continue training for another 100K steps at $224 \times 448$ resolution. The number of context views is sampled uniformly from $[2, 8]$ per iteration during multi-view training, and the context frame gap warms up from $[40, 50]$ to $[50, 200]$ frames on RE10K and from $[15, 30]$ to $[20, 50]$ on DL3DV. Unless otherwise noted, the learning rate is $5 \times 10^{-5}$ with a backbone multiplier of $0.01\times$ and batch size~1 per GPU. The fixed 6-view checkpoints use the same dataset-specific step budgets and resolution schedules, but keep the number of context views fixed to~6; for this setting, the learning rate is raised to $1 \times 10^{-4}$ and the batch size is increased to~4 across 4~GPUs. The YoNoSplat baseline follows the same training strategy as our model for controlled comparison.

\boldstart{Resolution and fair comparison.}
Following YoNoSplat~\citep{ye2025yonosplat}, we adopt its resolution and fair-comparison protocol for novel-view synthesis: we use the $224 \times 224$ version of our model because it best aligns with the experimental settings of the other baselines. Different prior methods adopt different input resolutions: MVSplat~\citep{chen2024mvsplat} and NoPoSplat~\citep{ye2024no} use $256 \times 256$, while DepthSplat~\citep{xu2024depthsplat} uses $256 \times 448$. Due to computational constraints and to avoid noise from in-house reproduction, it is not feasible to retrain every baseline and our model at a unified resolution. As in YoNoSplat, we keep comparisons fair in two ways. First, our $224 \times 224$ model has the smallest receptive size among the compared methods; because all methods first center-crop and then resize their inputs, square crops provide the most conservative receptive coverage. Second, since our model uses the smallest receptive size, rendered outputs from other methods can be center-cropped and resized so that all quantitative and qualitative comparisons are performed on the same image content.

\boldstart{Optimizer.}
We use AdamW with a linear warm-up of 2{,}000 steps and gradient clipping at~0.5.

\boldstart{Scheduled sampling.}
The probability of using predicted poses increases linearly from~0 to~0.9 between steps 160K and 200K during Stage~1 on RE10K.

\boldstart{Progressive sharpening schedules.}
The opacity exponent $e(t)$ ramps from $e_{\mathrm{init}} = 1$ to $e_{\mathrm{final}} = 2$ over the warm-up phase. The opacity temperature $\tau(t)$ ramps from $\tau_{\mathrm{init}} = 1.0$ to $\tau_{\mathrm{final}} = 5.0$ over $16{,}000$ steps. The blur multiplier $\beta(t)$ decays from $\beta_{\mathrm{init}} = 1.0$ to $\beta_{\mathrm{final}} = 0.5$ over $16{,}000$ steps.

\boldstart{Large-loss filtering.}
After 40K warm-up steps, training samples whose total loss exceeds~0.2 (or with MSE~$>$~0.06 or pose loss~$>$~1.0) have their loss contribution scaled to a negligible value.

\subsection{Mono-Normal Teacher}
\label{sec:app_mono_teacher}

The monocular normal teacher is the Omnidata DPT normal estimator~\citep{eftekhar2021omnidata} with a ViT-B/ResNet-50 hybrid backbone (variant ``vitb\_rn50\_384''). Teacher normals are computed offline for every input view and resized to match the prediction resolution via bilinear interpolation. The bootstrap schedule described in Sec.~\ref{sec:normal_pipeline} of the main paper uses $t_{\mathrm{tk}} = 6{,}000$ steps and $t_{\mathrm{bl}} = 20{,}000$ steps.

\section{Baseline Mesh Extraction: Quantitative Details}
\label{sec:app_baseline_mesh}

The main paper describes the TSDF fusion pipeline used for Gaussian baselines and the direct export pipeline of \method at a conceptual level. Here we report the exact numerical parameters used in both pipelines, which are necessary for reproducibility.

\boldstart{TSDF fusion parameters.}
The TSDF volume uses a voxel size of~0.005, SDF truncation of~0.1, and depth truncation of~5.0. Pixels with rendered alpha below~0.3 are masked out. During post-processing, connected component analysis retains the 50 largest clusters and removes clusters with fewer than~50 triangles.

\boldstart{Direct export parameters.}
The opacity threshold for triangle pruning is~0.10 after temperature scaling with $\tau$ set to the final training temperature of~5.0. Vertex deduplication uses quantized position hashing at precision~$10^{-5}$ with normal-octant keying to prevent merging across opposing face orientations. Per-triangle colors are computed from the 0th-order SH coefficients as $\mathbf{c} = \mathrm{clamp}(\mathrm{SH}_0 \cdot C_0 + 0.5,\, 0,\, 1)$ with $C_0 = \frac{1}{2\sqrt{\pi}} \approx 0.282$. The entire export completes in less than~0.1\,s on a single GPU, compared to more than~15\,s for TSDF fusion.

\section{Primitive Rendering Comparison}
\label{sec:app_prim_render}

\begin{figure}[!htbp]
\centering
\includegraphics[width=\textwidth]{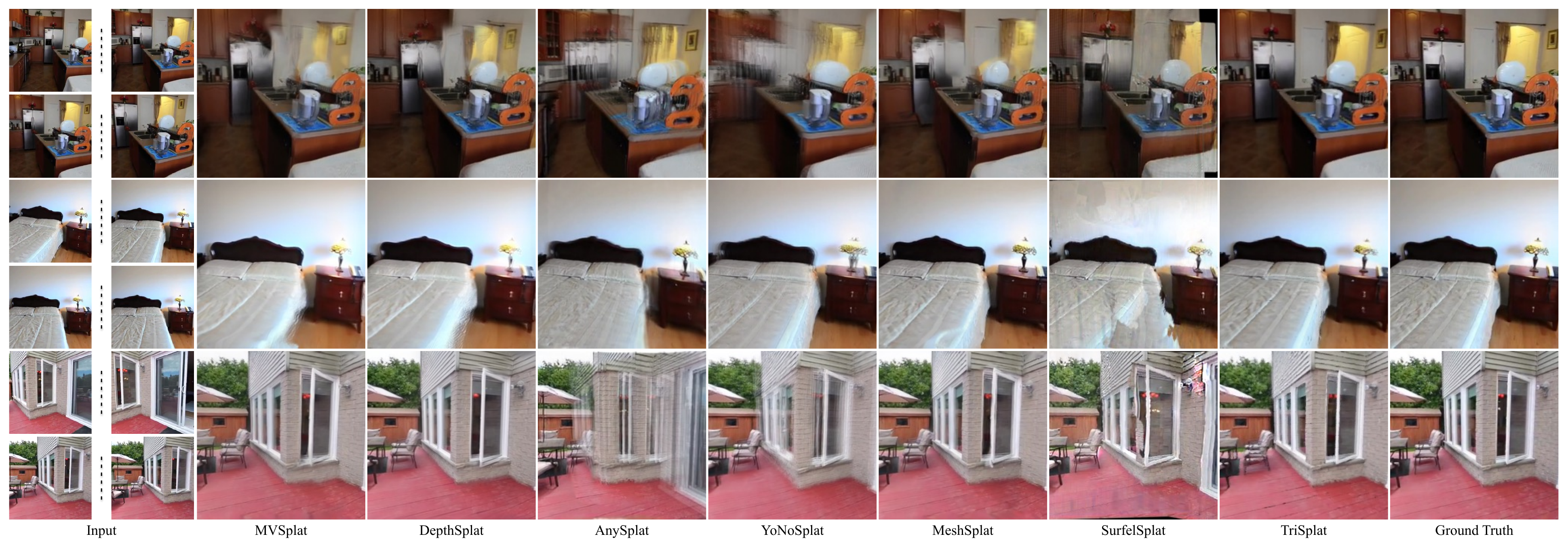}
\caption{\textbf{Primitive-rendering comparison on RE10K.} Each method is rendered with its own native rasterizer (Gaussian splatting for baselines, triangle splatting for \method). Compared with the mesh-rendering results in Fig.~\ref{fig:nvs_qual_mesh_re10k}, Gaussian baselines appear closer to \method under primitive rendering, but this quality does not transfer to the exported mesh used by downstream simulation and graphics pipelines.}
\label{fig:nvs_qual_prim}
\label{fig:nvs_qual_prim_re10k}
\end{figure}

The main paper reports rendering quality under \emph{mesh rendering}, which matches the simulation-ready objective of \method: a standard triangle rasterizer consumes the exported mesh directly. For completeness, we also report \emph{primitive rendering}, in which each method renders using its own native rasterizer prior to any mesh export (Gaussian splatting for baselines, triangle splatting for \method). Tables~\ref{tab:app_prim_dl3dv} and~\ref{tab:app_prim_re10k} report primitive-rendering metrics on DL3DV and RE10K, and Table~\ref{tab:app_prim_to_mesh} summarizes the primitive-to-mesh PSNR degradation for each method. Qualitative primitive-rendering comparisons are shown in Fig.~\ref{fig:nvs_qual_prim}.

\boldstart{Observations.}
Under primitive rendering, Gaussian baselines attain their strongest numerical scores because their smooth radial falloff provides locally forgiving gradient coverage at render time. When the same models are consumed as meshes, however, the TSDF-fusion step discards these smooth primitives and yields a substantially lower mesh-rendering PSNR (see Table~\ref{tab:app_prim_to_mesh}). \method exhibits a markedly smaller primitive-to-mesh degradation because the rendering primitives \emph{are} the exported triangles, so no information is lost during mesh construction. This property is central to the simulation-ready claim: the same representation used during training and inference can be consumed as a mesh with minimal quality loss, without relying on fragile post-hoc surface extraction.

\begin{figure}[!htbp]
\centering
\includegraphics[width=\textwidth]{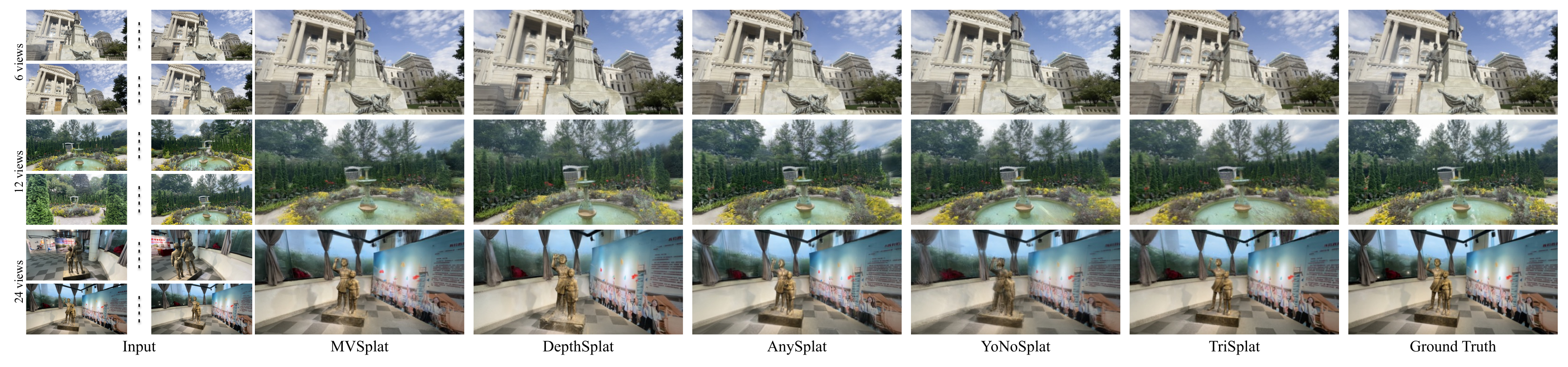}
\caption{\textbf{Primitive-rendering comparison on DL3DV.} Rows correspond to 6, 12, and 24 input views, and each method is rendered with its own native rasterizer before mesh export. These results provide the counterpart to Fig.~\ref{fig:nvs_qual_mesh_dl3dv}: Gaussian baselines can look substantially stronger before TSDF fusion, but their native rendering quality is not the representation consumed by standard mesh pipelines.}
\label{fig:nvs_qual_prim_dl3dv}
\end{figure}

\begin{table}[!htbp]
\centering
\caption{\textbf{Primitive rendering on DL3DV.} Metrics reported under each method's native rasterizer. Mesh-rendering results under the same evaluation protocol are reported in the main paper (Table~\ref{tab:main_dl3dv_nvs}).}
\label{tab:app_prim_dl3dv}
\resizebox{0.76\textwidth}{!}{
\begin{tabular}{l|ccc|ccc|ccc}
\toprule
& \multicolumn{3}{c|}{6 views} & \multicolumn{3}{c|}{12 views} & \multicolumn{3}{c}{24 views} \\
Method & PSNR$\uparrow$ & SSIM$\uparrow$ & LPIPS$\downarrow$ & PSNR$\uparrow$ & SSIM$\uparrow$ & LPIPS$\downarrow$ & PSNR$\uparrow$ & SSIM$\uparrow$ & LPIPS$\downarrow$ \\
\midrule
MVSplat      &  22.83 & 0.772 & 0.190 & 21.29 & 0.721 & 0.244   & 19.93 & 0.672 & 0.293 \\
DepthSplat   & 23.56 & 0.806 & 0.150 & 21.89 & 0.759 & 0.196 & 20.09 & 0.699 & 0.256 \\
AnySplat     & 19.79 & 0.579 & 0.257 & 19.95 & 0.587 & 0.272 & 20.21 & 0.608 & 0.274 \\
YoNoSplat    & 23.88 & 0.793 & 0.154 & 22.91 & 0.762 & 0.180 & 22.22 & 0.740 & 0.199 \\

\midrule
\method (Ours) & 23.22 & 0.768 &  0.218 & 21.87 & 0.726 & 0.260 & 21.16 & 0.704 &  0.282 \\
\bottomrule
\end{tabular}
}
\end{table}

\begin{table}[!htbp]
\centering
\caption{\textbf{Primitive rendering on RE10K (6 views).} Metrics are reported under each method's native rasterizer before mesh export. These numbers contextualize Table~\ref{tab:main_re10k}: Gaussian baselines are competitive under their original splatting renderers, but the downstream mesh-rendering protocol exposes the quality loss introduced by TSDF conversion.}
\label{tab:app_prim_re10k}
\resizebox{0.46\textwidth}{!}{
\begin{tabular}{l|ccc}
\toprule
Method & PSNR$\uparrow$ & SSIM$\uparrow$ & LPIPS$\downarrow$ \\
\midrule
MVSplat      & 24.27 & 0.839 & 0.146 \\
DepthSplat   & 25.32 & 0.849 & 0.139 \\
AnySplat     & 17.78      & 0.551      &     0.348   \\
YoNoSplat    & 27.19 & 0.889 & 0.106 \\
MeshSplat    & 22.90 & 0.788 & 0.188 \\
SurfelSplat  & 17.86 & 0.573 & 0.373 \\
\midrule
\method (Ours) &  26.46   &   0.870    &  0.130      \\
\bottomrule
\end{tabular}
}
\end{table}

\begin{table}[!htbp]
\centering
\caption{\textbf{Primitive$\to$Mesh degradation on RE10K (6 views).} We report PSNR under primitive rendering (Prim.) and mesh rendering (Mesh), and their difference $\Delta$. A smaller $|\Delta|$ indicates that the exported mesh faithfully preserves the rendering quality of the native primitives.}
\label{tab:app_prim_to_mesh}
\resizebox{0.48\textwidth}{!}{
\begin{tabular}{l|ccc}
\toprule
Method & Prim. PSNR & Mesh PSNR & $\Delta$PSNR \\
\midrule
MVSplat      & 24.27 & 13.57 & $-$10.70 \\
DepthSplat   & 25.32 & 19.65 & $-$5.67 \\
YoNoSplat    & 27.19 & 21.07 & $-$6.12 \\
MeshSplat    & 22.90 & 19.72 & $-$3.18 \\
SurfelSplat  & 17.86 &  9.09 & $-$8.77 \\
\midrule
\method (Ours) &  26.46  & 23.25 &  -3.21\\
\bottomrule
\end{tabular}
}
\end{table}

\section{Opacity Mapping Analysis}
\label{sec:app_opacity_derivation}

The opacity mapping in Eq.~\eqref{eq:opacity} satisfies four useful properties. First, boundary values are preserved for all $e > 0$ since $o(0;\,e) = 0$ and $o(1;\,e) = 1$, ensuring that fully transparent and fully opaque primitives remain unchanged regardless of the schedule. Second, when $e = 1$ the mapping reduces to identity ($o = p$), providing a natural starting point. Third, as $e$ increases, intermediate values of $p$ are pushed toward~0 or~1 so that $o(p;\,e) \to \mathbf{1}_{p > 0.5}$ in the limit $e \to \infty$, progressively binarizing the opacity field. Fourth, the mapping is differentiable everywhere in $(0,1)$ for $e > 0$, ensuring stable gradient flow.

The temperature factor described in Sec.~\ref{sec:progressive} increases linearly from $\tau_{\mathrm{init}} = 1.0$ to $\tau_{\mathrm{final}} = 5.0$ over 16{,}000 steps (experiment config overrides the decoder default of 8{,}000 steps). The dual mechanism, combining the exponent-based nonlinearity with temperature-driven sharpening, provides a richer curriculum than either component alone. An alpha floor of~0.02 is applied during early training to prevent premature pruning of uncertain primitives.

\section{Triangle Adapter Details}
\label{sec:app_triangle_adapter}

The main paper describes the triangle construction process at the formula level (Eq.~\eqref{eq:vertex}). Here we provide code-level details necessary for reproduction.

The canonical equilateral template uses three vertices:
\[
(0,\, 0.577,\, 0),\quad
(-0.5,\, -0.289,\, 0),\quad
(0.5,\, -0.289,\, 0).
\]
It is pre-scaled by a factor of~4. The three sigmoid-mapped scale logits are bounded to $[s_{\min},\, s_{\max}]$ and converted to world-space sizes using the predicted depth and a pixel-footprint multiplier derived from the inverse intrinsic matrix. During Stage~1 on RE10K the range is $[0.5, 18.0]$; during Stage~2 it is $[1.2, 15.0]$.

An optional coverage boosting mechanism increases the scale of low-confidence triangles. When the opacity falls below a threshold of~0.20, the scale is boosted proportionally to the gap between the threshold and the opacity, encouraging uncertain triangles to cover a wider area and receive more photometric gradients.

The blur parameter $\hat{\sigma}$ is converted to a positive value via $\sigma = \mathrm{sigmoid}(\hat{\sigma}) \cdot \beta(t) + \epsilon$, where $\beta(t)$ decays linearly from 1.0 to 0.5 over 16{,}000 steps.

At pixels where the geometry-based tangent-frame rotation is valid, the network's predicted quaternion is overridden by the geometry-derived quaternion. At invalid pixels (boundary pixels and degenerate cross products) the network quaternion is retained as a fallback.

\section{Additional Results on ScanNet}
\label{sec:supp_scannet_nvs}

Table~\ref{tab:nvs_scannet} reports novel-view synthesis results on ScanNet under the zero-shot setting, using models trained on RE10K without fine-tuning. Following the convention adopted in the main paper, we report mesh rendering as the primary metric, with primitive rendering included for reference. Despite the significant domain gap between real-estate walkthrough videos and indoor scans, \method maintains competitive performance under mesh rendering. The primitive-to-mesh degradation observed on the training datasets carries over to this unseen domain: Gaussian baselines lose a large margin of PSNR when their TSDF meshes are rendered as triangles, while \method remains stable, confirming that the direct-export property is a domain-robust effect rather than a dataset-specific artifact.

\begin{table}[!htbp]
\centering
\caption{\textbf{Novel-view synthesis on ScanNet (zero-shot, 6 views).} Models trained on RE10K, evaluated without fine-tuning. Primitive rendering (Prim.) is each method's native rasterization; mesh rendering (Mesh) rasterizes the exported mesh.}
\label{tab:nvs_scannet}
\resizebox{0.78\textwidth}{!}{
\begin{tabular}{l|ccc|ccc}
\toprule
& \multicolumn{3}{c|}{Prim.} & \multicolumn{3}{c}{Mesh} \\
Method & PSNR$\uparrow$ & SSIM$\uparrow$ & LPIPS$\downarrow$ & PSNR$\uparrow$ & SSIM$\uparrow$ & LPIPS$\downarrow$ \\
\midrule
MVSplat     & 17.64 & 0.639 & 0.345 & 11.72 & 0.320 & 0.625 \\
DepthSplat  & 17.15 & 0.589 & 0.438 & 16.06 & 0.534 & 0.519 \\
AnySplat    & 13.18 & 0.420 & 0.571 & 12.67 & 0.430 & 0.598\\
YoNoSplat   & 22.99 & 0.809 & 0.240 & 16.69 & 0.587 & 0.445 \\
MeshSplat   & 17.62 & 0.633 & 0.352 & 14.35 & 0.482 & 0.578 \\
SurfelSplat & 12.69 & 0.329 & 0.592 & 11.05 & 0.385 & 0.752 \\
\midrule
\method (Ours) & 22.61 & 0.796 & 0.265 & 17.03 & 0.540 & 0.530 \\
\bottomrule
\end{tabular}
}
\end{table}

\section{Additional Ablation Studies}
\label{sec:app_ablation}

The main-paper ablation (Table~\ref{tab:ablation}) validates the four key design choices at a coarse level. Here we provide finer-grained studies on hyperparameters and architectural variants. Consistent with the main paper, all tables in this section report surface geometry (CD, F1) together with mesh-rendering quality (PSNR, LPIPS), so that every design choice is evaluated on the same simulation-ready metrics used in the main experiments.

\subsection{Triangle Scale Range}
\label{sec:app_ablation_scale}

Table~\ref{tab:ablation_scale} examines the effect of the triangle scale range $[s_{\min}, s_{\max}]$. A range that is too narrow limits the model's ability to cover large surface regions, reducing recall. A range that is too wide permits excessively large triangles that introduce rendering artifacts.

\begin{table}[h]
\centering
\caption{\textbf{Ablation on triangle scale range} (RE10K, 6 views).}
\label{tab:ablation_scale}
\resizebox{0.62\textwidth}{!}{
\begin{tabular}{cc|cc|cc}
\toprule
$s_{\min}$ & $s_{\max}$ & CD$\downarrow$ & F1$\uparrow$ & PSNR$\uparrow$ & LPIPS$\downarrow$ \\
\midrule
0.5 & 10.0 & 0.189 & 0.619 &22.14  & 0.375 \\
0.5 & 18.0 & 0.190 & 0.708 & 23.25 & 0.318 \\
1.2 & 15.0 & 0.192 & 0.615 & 22.49 & 0.357 \\
1.2 & 25.0 & 0.181 & 0.631 & 19.97 & 0.452 \\
\bottomrule
\end{tabular}
}
\end{table}

\subsection{Blur Schedule}
\label{sec:app_ablation_blur}

Table~\ref{tab:ablation_blur} isolates the effect of blur scheduling. Without scheduling (fixed low blur) early training suffers from poor gradient coverage. A fixed high blur allows stable training but produces soft surfaces. The default schedule decaying from~1.0 to~0.5 over 16K steps achieves the best balance.

\begin{table}[h]
\centering
\caption{\textbf{Ablation on blur scheduling} (RE10K, 6 views).}
\label{tab:ablation_blur}
\resizebox{0.72\textwidth}{!}{
\begin{tabular}{l|cc|cc}
\toprule
Blur strategy & CD$\downarrow$ & F1$\uparrow$ & PSNR$\uparrow$ & LPIPS$\downarrow$ \\
\midrule
Fixed low ($\beta = 0.1$)     & 0.185 &0.640  & 21.44 &0.402   \\
Fixed high ($\beta = 1.0$)    & 0.223  & 0.542 & 20.90 &  0.434\\
Schedule $1.0 \to 0.5$ (8K)    &0.217   & 0.571 &  21.05&  0.418 \\
Schedule $1.0 \to 0.5$ (16K)   & 0.190 & 0.708 & 23.25 & 0.318 \\
Schedule $1.0 \to 0.1$ (16K)   & 0.212 &  0.588& 21.26  & 0.384 \\
\bottomrule
\end{tabular}
}
\end{table}

\subsection{Opacity Temperature}
\label{sec:app_ablation_temperature}

Table~\ref{tab:ablation_temperature} studies the opacity temperature schedule. Without temperature scaling ($\tau = 1.0$ throughout) the opacity distribution remains soft and the resulting semi-transparent surfaces degrade mesh quality. A very high final temperature ($\tau = 25.0$) produces near-binary opacities that cause gradient instability.

\begin{table}[h]
\centering
\caption{\textbf{Ablation on opacity temperature} (RE10K, 6 views).}
\label{tab:ablation_temperature}
\resizebox{0.68\textwidth}{!}{
\begin{tabular}{l|cc|cc}
\toprule
Temperature schedule & CD$\downarrow$ & F1$\uparrow$ & PSNR$\uparrow$ & LPIPS$\downarrow$ \\
\midrule
Fixed $\tau = 1.0$              &0.183  & 0.637 & 20.88 & 0.439 \\
$\tau\!\!: 1.0 \to 5.0$ (16K)     & 0.190 & 0.708 & 23.25 & 0.318  \\
$\tau\!\!: 1.0 \to 10.0$ (16K)   &  0.204 & 0.560 & 21.87 & 0.379 \\
$\tau\!\!: 1.0 \to 25.0$ (16K)   &0.232  &  0.454&  17.23& 0.477 \\
\bottomrule
\end{tabular}
}
\end{table}

\section{Additional Simulation Experiments}
\label{sec:app_simulation}

The main paper summarizes robotic grasping, ball dynamics, and multi-platform locomotion in Unity and NVIDIA Isaac Sim (Fig.~\ref{fig:sim_demo}). Here we expand the simulation demonstrations into four-frame dynamic sequences using the directly exported triangle meshes without any manual cleanup or format conversion. Frames are ordered from left to right by time.

\subsection{Rigid-Body Dynamics}
\label{sec:app_rigid_body}
We evaluate the physical utility of our reconstructed meshes through two rigid-body scenarios simulated in NVIDIA Isaac Sim with the PhysX backend. First, in a ball drop experiment, a sphere is released from various heights onto the surface. The resulting collision responses--including intricate bounce trajectories--demonstrate that our mesh faithfully captures the underlying geometry with high fidelity. Second, in an object stacking experiment, we place multiple rigid objects on the reconstructed surfaces. The sustained stability of these stacks highlights the exceptional surface flatness and normal consistency achieved by our method, ensuring reliable contact physics for downstream interaction tasks.

\subsection{Legged Locomotion}
\label{sec:app_locomotion}

A simulated quadruped robot traverses outdoor scenes reconstructed by \method. The robot's foothold planning directly leverages the mesh surface normals and collision geometry. In scenes containing stairs and chairs, the robot successfully navigates the terrain.

\begin{figure}[!htbp]
\centering
\includegraphics[width=\textwidth]{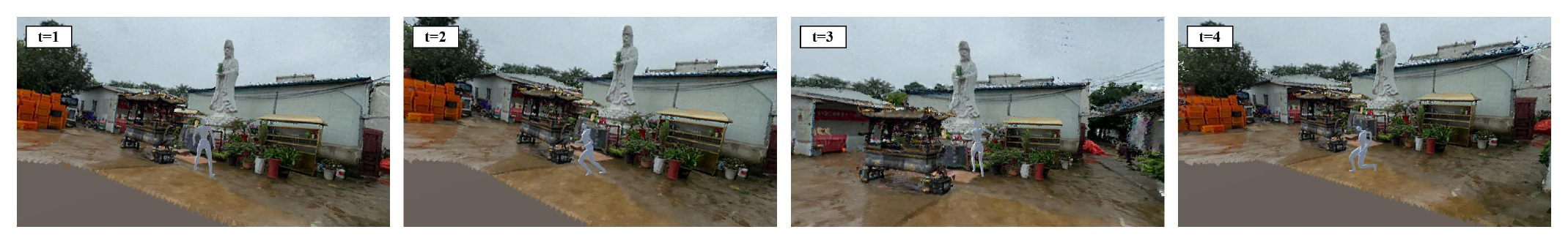}
\caption{\textbf{Unity character navigation.} The exported \method mesh is imported into Unity as static scene geometry for character navigation and collision handling. The four frames show temporal progression from $t=1$ to $t=4$ and demonstrate usable collision surfaces without manual cleanup, conversion from Gaussian primitives, or scene-specific mesh repair.}
\label{fig:supp_sim_unity_character}
\end{figure}

\begin{figure}[!htbp]
\centering
\includegraphics[width=\textwidth]{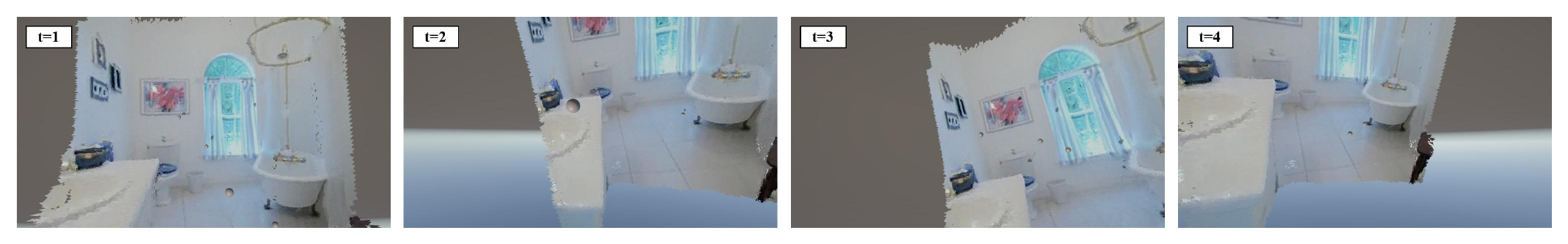}
\caption{\textbf{Unity object interaction.} We load the reconstructed mesh into Unity and interact with the scene using standard engine collision and physics components. The four frames show temporal progression from $t=1$ to $t=4$; stable contact with reconstructed surfaces indicates that the exported triangle mesh is directly usable for interactive applications, rather than only for image rendering.}
\label{fig:supp_sim_unity_interaction}
\end{figure}

\begin{figure}[!htbp]
\centering
\includegraphics[width=\textwidth]{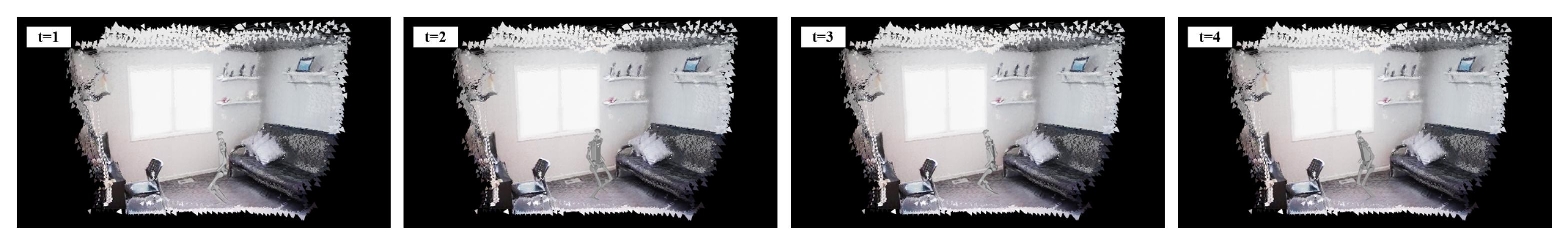}
\caption{\textbf{Isaac Sim humanoid locomotion.} The four frames show the H1 humanoid moving over the same exported \method mesh from $t=1$ to $t=4$. This sequence expands the compact main-paper demonstration by showing continuous contact-rich locomotion on the reconstructed triangle geometry, without an intermediate reconstruction or mesh-repair stage.}
\label{fig:supp_sim_isaac_h1}
\end{figure}

\begin{figure}[!htbp]
\centering
\includegraphics[width=\textwidth]{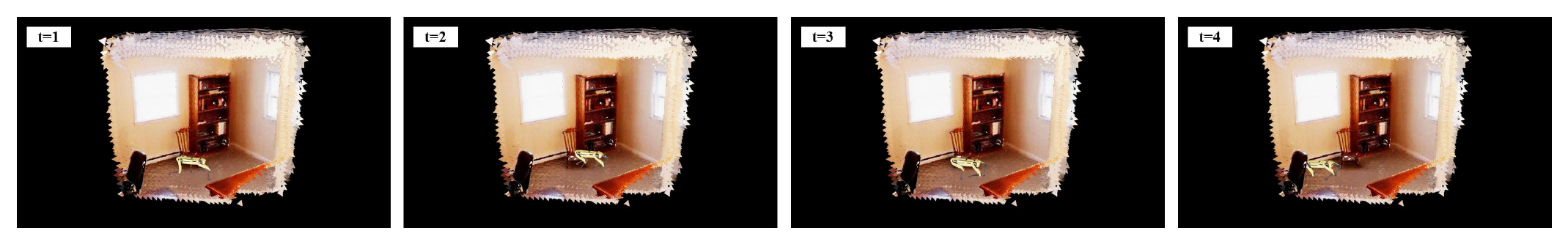}
\caption{\textbf{Isaac Sim quadruped locomotion.} A quadruped traverses the exported \method mesh using the simulator's native collision and contact solver. The four frames show temporal progression from $t=1$ to $t=4$ and illustrate that the same mesh representation supports different embodied agents and locomotion patterns.}
\label{fig:supp_sim_isaac_quad}
\end{figure}

\section{More Visual Comparisons}
\label{sec:app_visual}

\begin{figure}[!htbp]
\centering
\includegraphics[width=\textwidth]{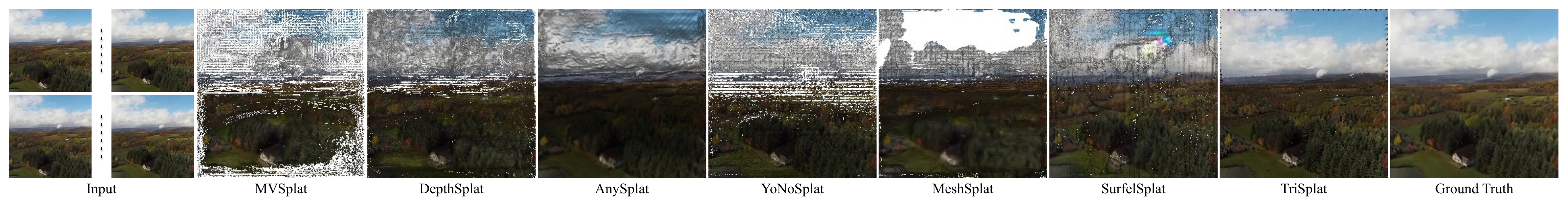}
\vspace{0.10cm}
\includegraphics[width=\textwidth]{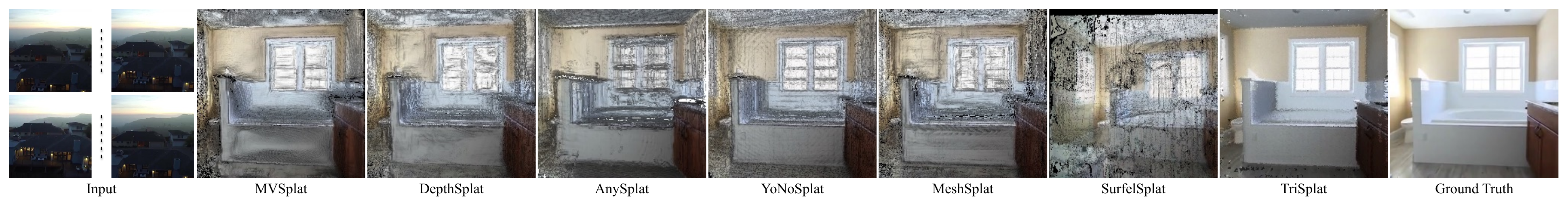}
\vspace{0.10cm}
\includegraphics[width=\textwidth]{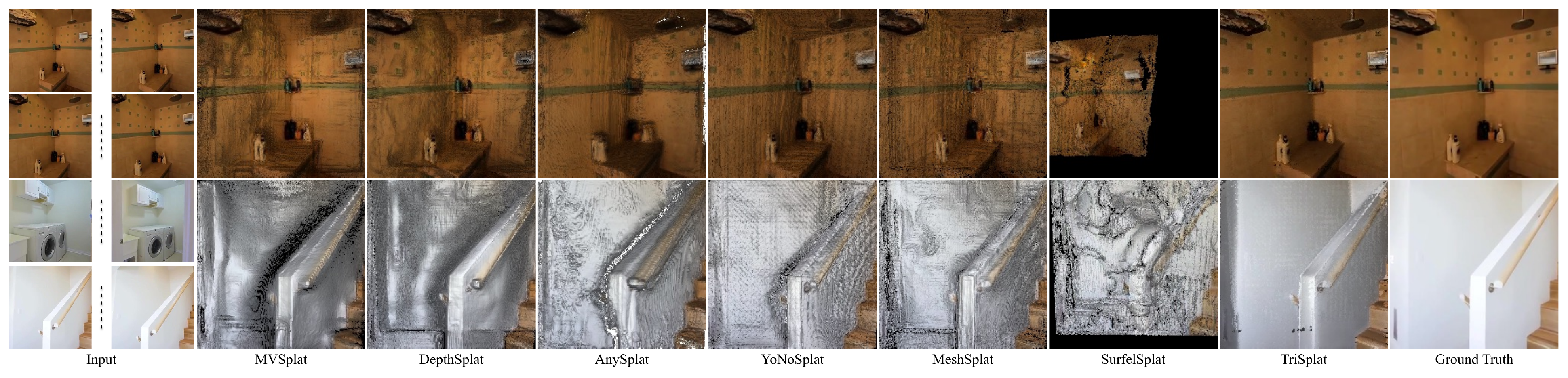}
\caption{\textbf{Additional mesh-rendering comparisons on RE10K.} We group all remaining RE10K mesh-rendering examples under one figure to avoid splitting identical comparisons across separate captions. Each group contains six input views, mesh-rendered baseline results, \method, and the ground-truth target view. Since every image is rendered from the exported mesh, artifacts reflect the quality of the representation available to downstream graphics and simulation systems.}
\label{fig:supp_nvs_re10k_all}
\label{fig:supp_nvs_re10k_01}
\label{fig:supp_nvs_re10k_02}
\label{fig:supp_nvs_re10k_03}
\end{figure}

\begin{figure}[!htbp]
\centering
\includegraphics[width=\textwidth]{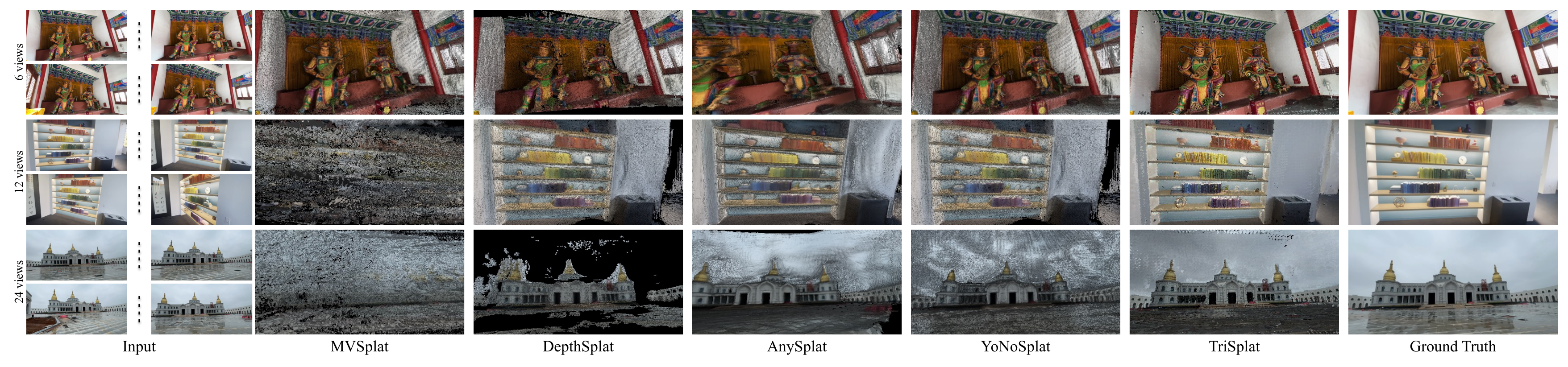}
\caption{\textbf{Additional mesh-rendering comparisons on DL3DV.} These examples are not used in the main-paper visual page. The comparison again isolates mesh-rendering quality: all methods are evaluated after export, so sharper results indicate a mesh that better preserves the original image evidence.}
\label{fig:supp_nvs_dl3dv_02}
\end{figure}

\begin{figure}[!htbp]
\centering
\includegraphics[width=\textwidth]{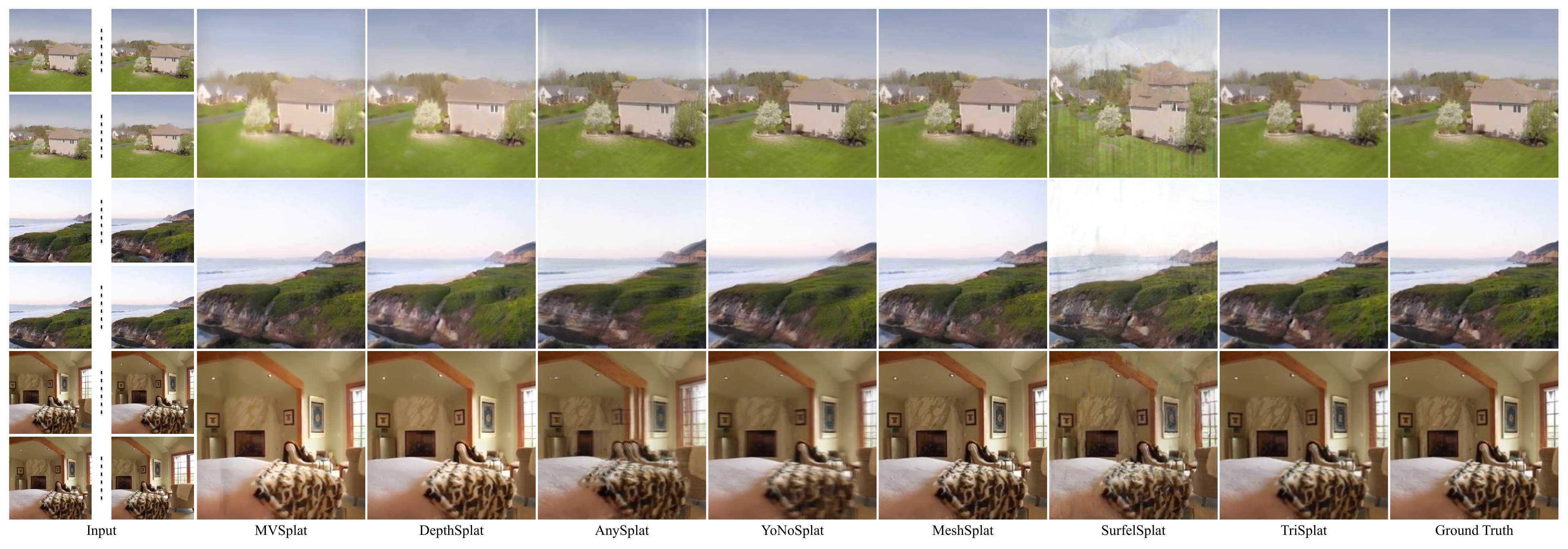}
\caption{\textbf{Additional primitive-rendering comparisons on RE10K.} Each method is rendered under its own native rasterizer before mesh export. These examples complement Fig.~\ref{fig:supp_nvs_re10k_all} by showing that strong Gaussian primitive renderings do not necessarily translate into high-quality exported meshes.}
\label{fig:supp_prim_re10k_02}
\end{figure}

\begin{figure}[!htbp]
\centering
\includegraphics[width=\textwidth]{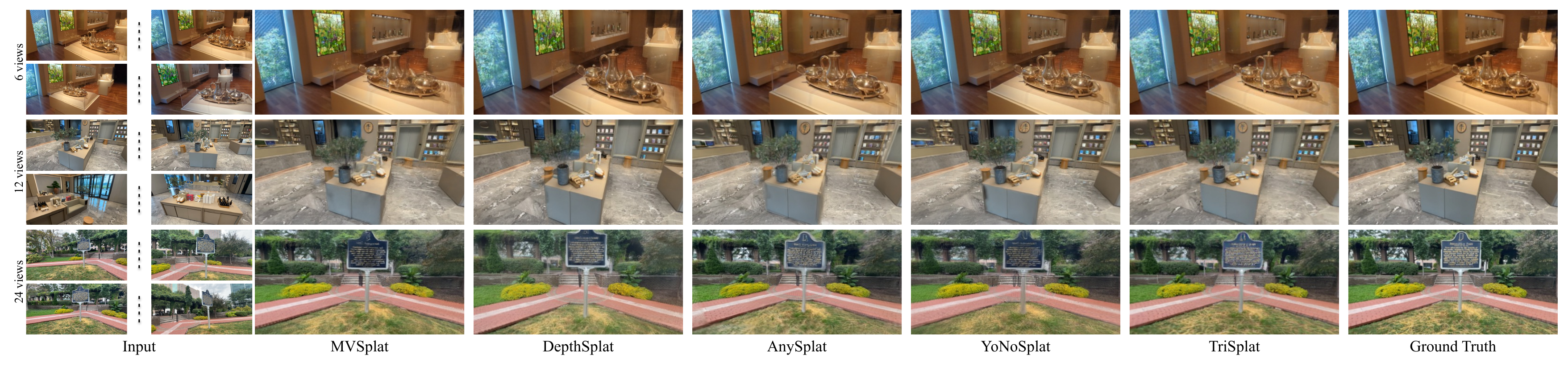}
\caption{\textbf{Additional primitive-rendering comparisons on DL3DV.} Native primitive renderings are shown for the same evaluation style as the DL3DV mesh-rendering figures. The contrast with Fig.~\ref{fig:supp_nvs_dl3dv_02} illustrates why mesh rendering is the relevant protocol for simulation-ready reconstruction.}
\label{fig:supp_prim_dl3dv_02}
\end{figure}

\begin{figure}[!htbp]
\centering
\includegraphics[width=\textwidth]{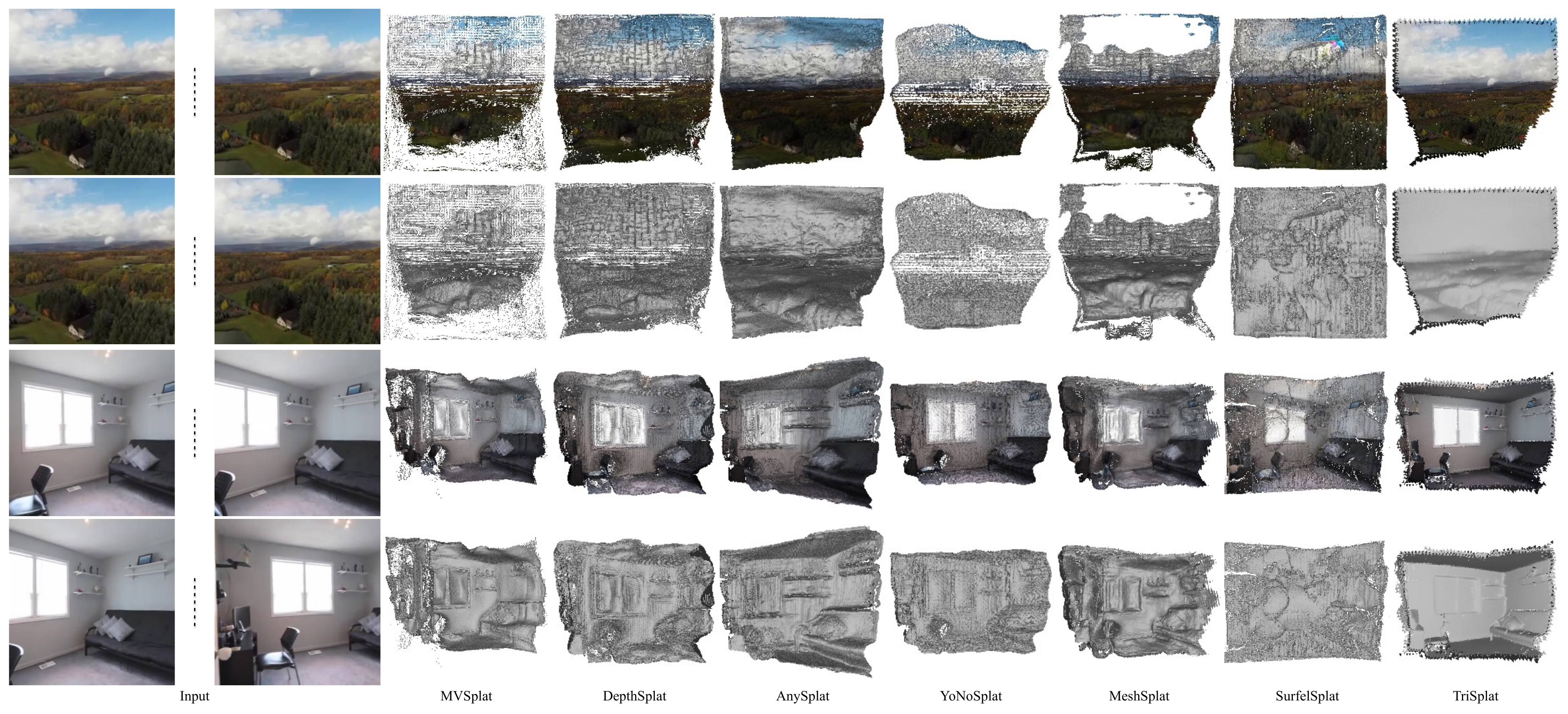}
\caption{\textbf{Additional textured mesh comparisons on RE10K.} The figure visualizes the remaining exported textured meshes rather than target-view renders. TSDF-fused Gaussian meshes often show over-smoothed surfaces, missing thin structures, and fragmented regions, while \method preserves direct triangle geometry and appearance in the exported representation.}
\label{fig:supp_mesh_re10k_01}
\end{figure}

\begin{figure}[!htbp]
\centering
\includegraphics[width=\textwidth]{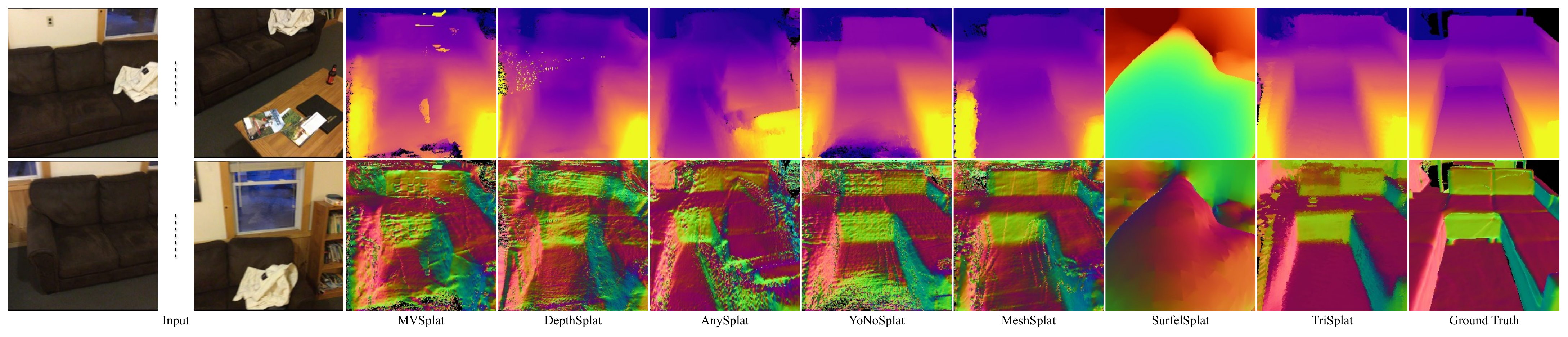}
\caption{\textbf{Additional depth and normal comparisons on ScanNet.} This example is not used in the main-paper visual page. All methods are trained on RE10K and evaluated on ScanNet without fine-tuning; \method produces smoother, geometrically coherent normals aligned with surface boundaries, while Gaussian baselines often produce noisy orientation fields under the domain shift.}
\label{fig:supp_scannet_depth_normal_01}
\end{figure}

\begin{figure}[!htbp]
\centering
\includegraphics[width=\textwidth]{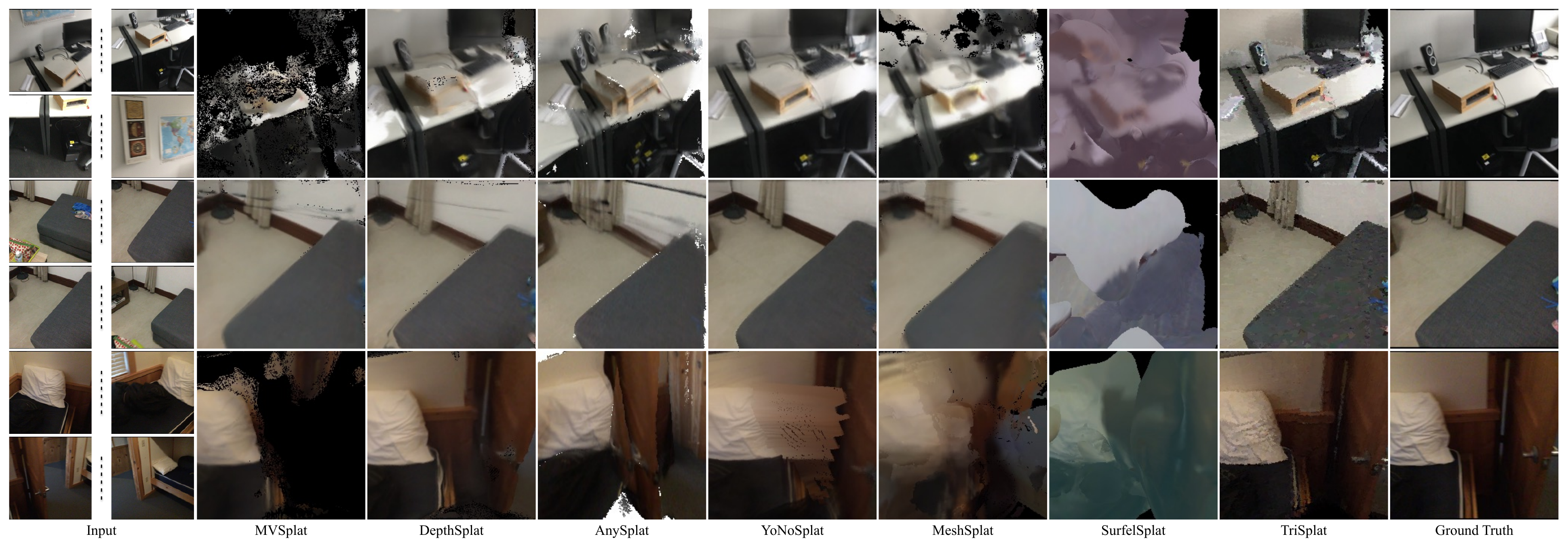}
\caption{\textbf{Primitive-rendering comparison on ScanNet.} Each method is rendered with its native primitive representation on zero-shot ScanNet scenes. These fixed-resolution renders are inserted without border cropping, preserving each method's original output frame and avoiding artificial enlargement or truncation of any baseline result.}
\label{fig:supp_scannet_prim_render}
\end{figure}

\begin{figure}[!htbp]
\centering
\includegraphics[width=\textwidth]{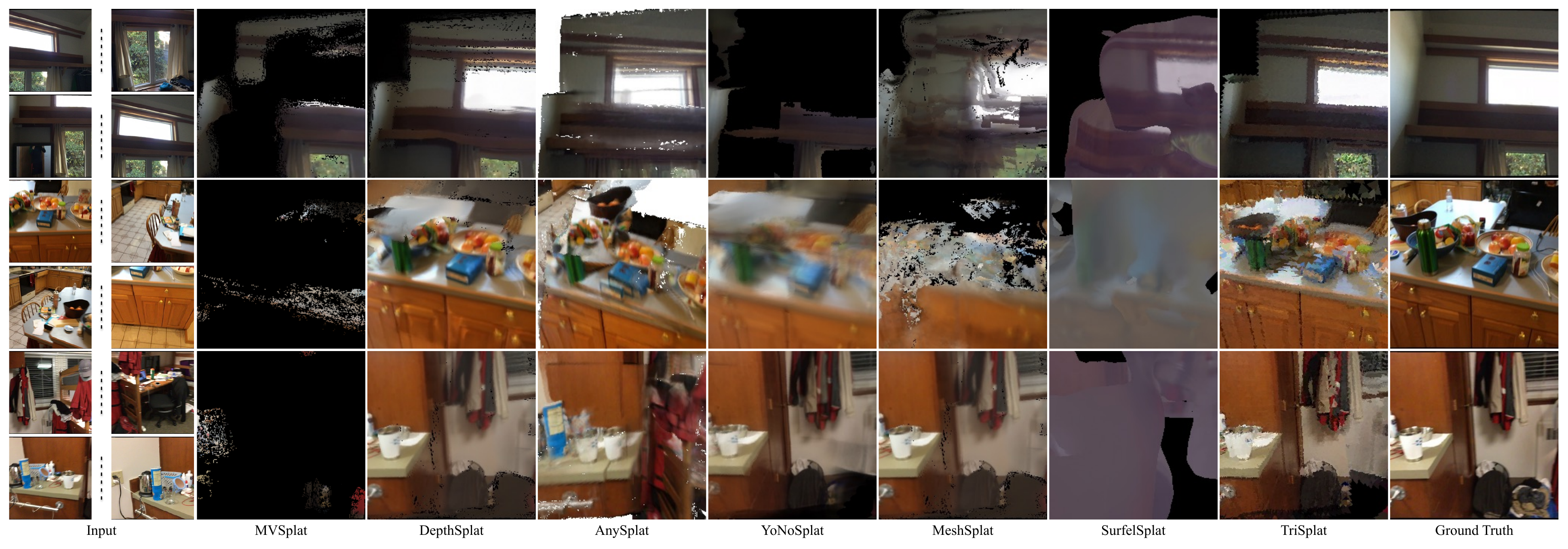}
\caption{\textbf{Mesh-rendering comparison on ScanNet.} We render each exported mesh with the same triangle pipeline in the zero-shot setting. Fixed-resolution outputs are inserted without border cropping, so the comparison preserves the full rendered frame for every baseline, including methods whose exported meshes contain large empty regions or incomplete surfaces.}
\label{fig:supp_scannet_mesh_render}
\end{figure}

We provide additional qualitative comparisons on RE10K~\citep{zhou2018stereo}, DL3DV~\cite{ling2024dl3dv}, and ScanNet~\cite{dai2017scannet}. These figures exclude scenes already shown in the main-paper visual pages. Fig.~\ref{fig:supp_nvs_re10k_all} groups the remaining RE10K mesh-rendering comparisons under one caption, Fig.~\ref{fig:supp_nvs_dl3dv_02} shows additional DL3DV mesh-rendering examples, and Figs.~\ref{fig:supp_prim_re10k_02} and~\ref{fig:supp_prim_dl3dv_02} show additional primitive-rendering examples. Figs.~\ref{fig:supp_mesh_re10k_01} and~\ref{fig:supp_scannet_depth_normal_01} show remaining textured-mesh and depth/normal examples, and Figs.~\ref{fig:supp_scannet_prim_render}--\ref{fig:supp_scannet_mesh_render} show zero-shot ScanNet rendering results.

\section{Mesh Evaluation Protocol}
\label{sec:app_mesh_eval}

The mesh metrics reported in the main paper (Tables~\ref{tab:main_dl3dv_surface} and~\ref{tab:main_re10k}) follow the standard protocol of MeshSplat~\citep{chang2025meshsplat} and ScanNet~\citep{dai2017scannet}. Both the predicted and ground-truth meshes are sampled into point clouds and voxel-downsampled at resolution~0.02 to ensure uniform density. One-sided distances are computed as $d(A, B) = \frac{1}{|A|} \sum_{\mathbf{a} \in A} \min_{\mathbf{b} \in B} \|\mathbf{a} - \mathbf{b}\|$, yielding Chamfer Distance $\mathrm{CD} = d(\mathcal{M}_p, \mathcal{M}_g) + d(\mathcal{M}_g, \mathcal{M}_p)$. Precision and Recall are the fractions of predicted and ground-truth points within distance $\delta = 0.05$ of each other, and the F1 score is their harmonic mean. All nearest-neighbor queries use KD-trees (Open3D) and all metrics are computed in the world coordinate frame.

\end{document}